\documentclass[10pt,twocolumn,letterpaper]{article}

\usepackage[pagenumbers]{cvpr} %

\usepackage{enumitem}

\usepackage[dvipsnames]{xcolor}

\usepackage{color}
\usepackage{multirow}

\usepackage{bm}
\usepackage{siunitx}
\usepackage{url}
\usepackage{booktabs}
\usepackage{tabularx}
\usepackage{adjustbox}
\usepackage{colortbl}

\usepackage{makecell}
\usepackage{soul}

\definecolor{turquoise}{cmyk}{0.65,0,0.1,0.3}
\definecolor{purple}{rgb}{0.65,0,0.65}
\definecolor{dark_green}{rgb}{0, 0.5, 0}
\definecolor{orange}{rgb}{0.8, 0.6, 0.2}
\definecolor{red}{rgb}{0.8, 0.2, 0.2}
\definecolor{darkred}{rgb}{0.6, 0.1, 0.05}
\definecolor{blueish}{rgb}{0.0, 0.3, .6}
\definecolor{light_gray}{rgb}{0.7, 0.7, .7}
\definecolor{pink}{rgb}{1, 0, 1}
\definecolor{greyblue}{rgb}{0.25, 0.25, 1}

\newcommand{\rgbgray}{\mathcal{G}}

\newcommand{\evimo}{EVIMOv2~\cite{EVIMO2}\xspace}

\newcommand{\evimosceneA}{depth\_var\_1\_lr\_000000\xspace}
\newcommand{\evimosceneB}{scene7\_00\_000001\xspace}
\newcommand{\evimosceneC}{scene8\_01\_000000\xspace}

\definecolor{color1}{rgb}{0.9, 0.65, 0.65}
\definecolor{color2}{rgb}{0.95, 0.8, 0.8}
\definecolor{color3}{rgb}{1.0, 0.9, 0.9}
\newcommand{\best}[1]{\hspace{-\fboxsep}\colorbox{color1}{\textbf{#1}}\hspace{-\fboxsep}}
\newcommand{\second}[1]{\hspace{-\fboxsep}\colorbox{color2}{\textit{#1}}\hspace{-\fboxsep}}

\def \customparskip {0.7em}
\renewcommand{\paragraph}[1]{\vspace{\customparskip}\noindent\textbf{#1.}}

\definecolor{cvprblue}{rgb}{0.21,0.49,0.74}
\usepackage[pagebackref,breaklinks,colorlinks,citecolor=cvprblue]{hyperref}
\usepackage{xcolor}

\def\confName{3DV\xspace}
\def\confYear{2025\xspace}

\title{LSE-NeRF: \ul{L}earning \ul{S}ensor Modeling \ul{E}rrors for Deblured Neural Radiance Fields with RGB-Event Stereo}

\author{
Wei Zhi Tang$^{1}$ \quad
Daniel Rebain$^{1}$ \quad
Konstantinos G. Derpanis$^{2}$ \quad
Kwang Moo Yi$^{1}$ \\
\vspace{0.002em} \\  % Adjust the value (0.5em) to increase or decrease the gap
$^{1}$University of British Columbia \\
$^{2}$York University \\
{\tt\small weiztang@cs.ubc.ca, drebain@cs.ubc.ca, kosta@yorku.ca, kmyi@cs.ubc.ca}
}

\begin{document}
\maketitle
\begin{abstract}

We present a method for reconstructing a clear Neural Radiance Field (NeRF) even with fast camera motions. 
To address blur artifacts, we leverage both (blurry) RGB images and event camera data captured in a binocular configuration. 
Importantly, when reconstructing our clear NeRF, we consider the camera modeling imperfections that arise from the simple pinhole camera model as learned embeddings for each camera measurement, and further learn a mapper that connects event camera measurements with RGB data.
As no previous dataset exists for our binocular setting, we introduce an event camera dataset with captures from a 3D-printed stereo configuration between RGB and event cameras.
Empirically, we evaluate our introduced dataset and EVIMOv2 and show that our method leads to improved reconstructions.
Our code and dataset are available at \href{https://github.com/ubc-vision/LSENeRF}{https://github.com/ubc-vision/LSENeRF}.

\end{abstract}

% We present a method for reconstructing a clear Neural Radiance Field (NeRF) even with fast camera motions. To address blur artifacts, we leverage both (blurry) RGB images and event camera data captured in a binocular configuration. Importantly, when reconstructing our clear NeRF, we consider the camera modeling imperfections that arise from the simple pinhole camera model as learned embeddings for each camera measurement, and further learn a mapper that connects event camera measurements with RGB data. As no previous dataset exists for our binocular setting, we introduce an event camera dataset with captures from a 3D-printed stereo configuration between RGB and event cameras. Empirically, we evaluate our introduced dataset and EVIMOv2 and show that our method leads to improved reconstructions. Our code and dataset are available at https://github.com/ubc-vision/LSENeRF.
    
\section{Introduction}
\label{sec:intro}

\begin{figure}
  \centering
  \includegraphics[width=0.48\textwidth]{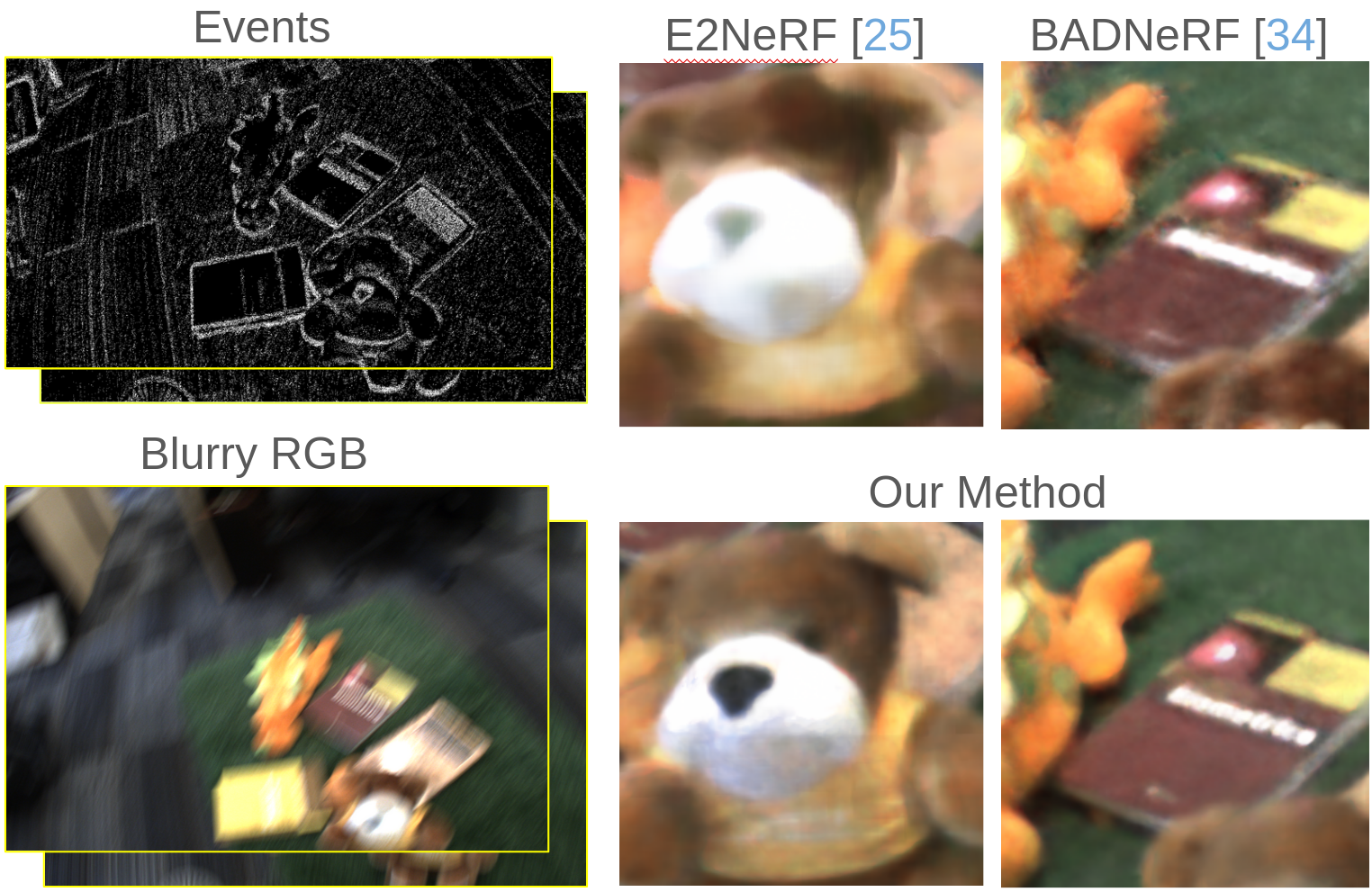}
  \caption{
  {\bf Teaser -- }
  We propose a deblur NeRF method that uses both RGB and event data. 
  We focus on sensor modeling imperfections, which allows our method to effectively make use of both modalities.
  As shown, our method provides significantly sharper reconstructions compared to both when using only RGB and also other RGB/event NeRF baselines.
  }
  \label{fig:teaser}
\end{figure}

Novel view synthesis has rapidly advanced since the recent advent of Neural Radiance Fields~(NeRFs)~\cite{nerf}.  
NeRFs learn a Multi-Layer Perceptron (MLP) to represent a scene, and use volume rendering to realize an image at a given camera pose.
Various extensions have been 
explored, including improvements to its efficiency~\cite{ingp}, robustness to occluders and transients~\cite{robustnerf}, tolerance to inexact input camera poses~\cite{barf, BADNeRF}, and applications to human-centric modeling~\cite{neuman}. 
More recently, 3D Gaussian splatting~\cite{gaussian-splat} has been proposed as an alternative representation based on rasterization instead of volume rendering.
Whether using NeRFs or Gaussians splats, building a scene representation capable of rendering high-quality novel views depends on having clear and sharp images during training.

Consequently, there has been focused research on removing this constraint
by augmenting existing neural rendering pipelines with a blur model, allowing for the acquisition of a deblurred reconstruction ~\cite{deblurnerf, BADNeRF, dpnerf}.
These works incorporate a physical model of blur formation under fast camera motions, and by doing so recover a non-blurred scene, as an inverse problem.
While these methods deblur scenes to some extent, under large camera motions they still suffer from imperfect reconstructions; see~\Cref{fig:teaser}. 
This is inevitable as the blur removes details from the original inputs, and there is only so much that can be recovered without additional priors.

To further mitigate this issue, researchers have also sought additional data modalities, specifically event cameras that can complement blurry RGB images~\cite{enerf, e2nerf, ebad-nerf,evDeblurNeRF}.
Event streams, unlike typical RGB images, do not suffer from motion blur~\cite{event-review}, hence incorporating them into NeRF pipelines can help deblur the scene.
Although these methods improve the clarity of NeRF reconstructions, they often focus on synthetic data~\cite{enerf,e2nerf,evDeblurNeRF,ebad-nerf} and are limited to single-camera scenarios that have aligned RGB and event data~\cite{e2nerf,evDeblurNeRF,ebad-nerf}. The latter restricts the type of devices that can be used, and the resolution remains relatively low ($640\times 480$)~\cite{denerf, e2nerf, ebad-nerf}.
In fact, for our target binocular setting, methods like E2NeRF~\cite{e2nerf} perform poorly, as demonstrated in \Cref{fig:teaser} and later in our experiments. Additionally, some methods, such as EvDeblurNeRF~\cite{evDeblurNeRF}, are entirely unsuitable because their loss function requires precise alignment between the RGB and event sensors.

In this work, we aim to better utilize RGB and event data to achieve deblurred NeRFs by also focusing on improved sensor modeling.
Specifically, (i) to model the sensor response differences between RGB and event data, we %
use a power mapping function, that is, the gamma function.
(ii) to take into account the per-measure variations that may happen due to various camera hardware functions, we utilize per-time embeddings. 
The former, learning a gamma mapping, is performed together with the NeRF training process, in contrast to the conventional constant threshold adaptation~\cite{e2nerf,ebad-nerf} and the normalization strategy~\cite{ev-rgbd-neural-slam, enerf}. 
This is similar in spirit to EvDeblurNeRF~\cite{evDeblurNeRF}, where the response functions are learned as MLPs, but as we %
show empirically, our solution, despite its simplicity, is superior.
For the latter, while per-time embeddings is a common strategy used in 
training conventional
NeRFs~\cite{nerfstudio} in the non-blurred case, we find that learning these embeddings, and then substituting them with a \emph{global} embedding that works well for all frames, is highly effective in mitigating blur---in fact, up to a level where it can outperform RGB/event NeRF baselines.

To validate our work, as no high-resolution binocular event-RGB dataset exists for deblurring applications that cover both indoor and outdoor environments, we introduce a new dataset with RGB images and corresponding events.
Specifically, we 3D print a stereo casing that holds a GigE Blackfly S RGB camera \cite{flir_blackfly_s_2024} and a Prophesee EVK-3 HD event camera \cite{evk3-hd}. 
We then capture five outdoor scenes and five indoor scenes that exhibit fast and slow camera motion for training and testing, respectively.
Unlike existing datasets that are captured with a single camera that provides both temporal and pixel \emph{aligned}
RGB and event streams, our dataset is binocular, with each camera providing high spatial resolution data of each modality. 
In more detail, our dataset provides $1440 \times 1080$ RGB images and event streams of resolution $1280 \times 720$, whereas existing datasets typically offer a resolution of $346 \times 260$ for both RGB and events.
We believe our dataset will be helpful when experimenting with systems that have both sensor modalities, installed in different physical locations, for example on augmented reality headsets or vehicles.
To facilitate research in this area, we will release our code, dataset
and 3D printing schematics.

To summarize, our contributions are as follow:
\begin{itemize}
    \item we introduce a novel method focused on sensor modeling errors for RGB and event-based deblur NeRF;
    \item to facilitate training and evaluation in our novel setting, we introduce a new dataset that provides high-resolution RGB and event streams in a binocular setup; and
    \item we significantly outperform the state of the art. %
\end{itemize}

\section{Related Work}

We first discuss previous work that focus on building a clear 3D representation from either (blurry) RGB images, event data, or a combination of both. Next, we briefly discuss RGB and event stream data used for deblur NeRF.

\paragraph{Deblur NeRF}
Since the introduction of NeRF~\cite{nerf}, many extensions have been developed to enhance the recovery of clear 3D neural representations from blurry images.
Deblur-NeRF~\cite{deblurnerf}  attempts to recover a clear 3D representation by explicitly modeling the blurring process as an averaging over multiple rays that potentially caused the blur, and optimizing a clear NeRF that would have created the blurry images.
More recent works~\cite{bad-gaussian, BADNeRF, exblurf} have observed that camera poses from blurry images are inherently inaccurate and further include camera optimization in the training process.
Other works further enforce rigid camera motion~\cite{dpnerf} or attempt to learn the 2D blur kernels with sharpness priors~\cite{sharp-nerf}.
In our work, we show that simply learning a per-time embedding during NeRF training, then substituting these embeddings with a \emph{global} scene embedding also works well for further deblurring, outperforming BADNeRF~\cite{BADNeRF}.

\paragraph{Event-based NeRF}
To leverage the high temporal resolution provided by event cameras, recent works have explored using event streams, with or without RGB images, to create a NeRF.
Concurrently, E-NeRF~\cite{enerf} and EventNeRF~\cite{col-enerf} are the first to explore the use of events and color events for neural fields, respectively. 
Later, methods that combine RGB and events to deblur neural fields appeared~\cite{e2nerf, evDeblurNeRF}.
The camera optimization idea of deblur NeRFs has also been extended to events and RGB NeRF~\cite{ebad-nerf}.
Other works also look into using RGB-D data in conjunction, for both static~\cite{ev-rgbd-neural-slam} and dynamic scenes~\cite{denerf}.

Besides NeRF, as 3D Gaussian splatting~\cite{gaussian-splat} has gained popularity, event-based methods have also been converted to Gaussian splats~\cite{eadeblur-gs, evagaussians}.
These include those~\cite{eadeblur-gs} that extend the event double integral (EDI)~\cite{edi} model used in event-based NeRFs~\cite{ebad-nerf}, or those~\cite{event3dgs} that include event sampling strategies to address the accumulation period question for frame-based event representation.

\paragraph{RGB and event stream data}
In all of the RGB and event works discussed above, except for DE-NeRF~\cite{denerf}, all real scenes are captured using a variant of the DAVIS~346~\cite{davis} camera---a camera that captures both RGB and event streams, at a low resolution of $346 \times 260$, and furthermore, provide data where the two modalities are physically aligned. 
This is a strong assumption, which limits the space of device setups should one build a system that utilizes both RGB and events.
Consequently, some work \emph{require} this setup as an assumption in their loss formulations~\cite{evDeblurNeRF} that limits their applicability.
In the case of DE-NeRF~\cite{denerf}, they also utilize \evimo, another dataset that provides both, but \emph{decoupled},  
RGB and event streams, but this dataset is limited to indoor scenarios, has artificial texture in some scenes, and the event resolution is still lower than ours---$640 \times 480$ whereas ours is $1280 \times 720$.
Further, as the intent of this dataset was event-based object segmentation, optical flow, and structure-from-motion, not all scenes are useful for evaluating deblur performance.
We thus collect our own high-resolution dataset, with a binocular setup that 
targets the deblurring task.

\section{Method}

\begin{figure}
    \centering
    \includegraphics[width=0.48\textwidth]{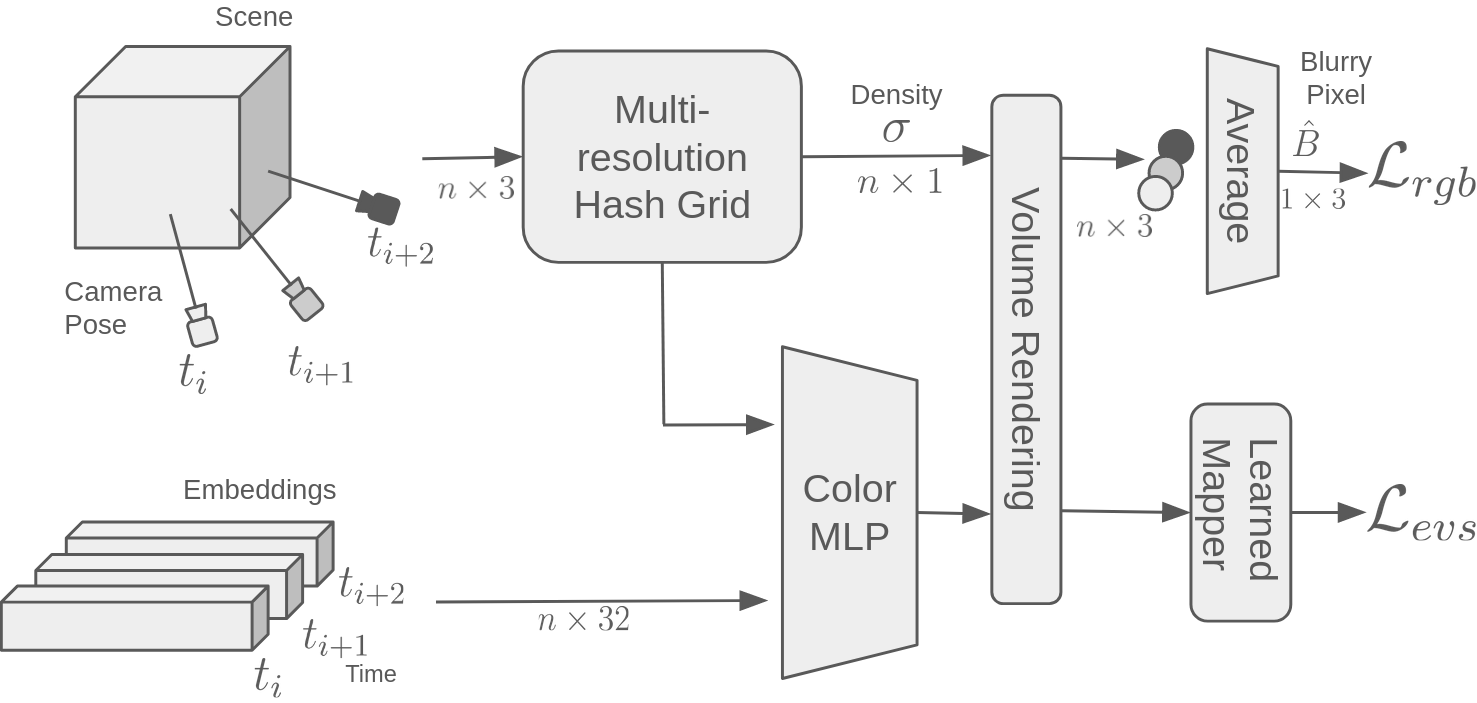}
      \caption{
      {\bf Framework overview -- }
      To render a pixel considering the camera motion blur, we pass points along the light rays of $n$ cameras through the hash grid to obtain their density and colors.
      We then volume render the pixel colors for the $n$ cameras and average them to generate the motion-blurred pixel color.
      We further utilize a learned mapper that maps RGB volume render to an intensity response for the event stream, which then utilizes the $n-1$ subsequent camera pairs to measure events.
      Note that our color Multi-Layer Perceptron (MLP) takes per-time learnable embeddings as input, to account for the sensor modeling imperfections.
      }
      \label{fig:framework}
  \end{figure}

Our main technical contribution are two-fold: (i) the learned per-time embedding that we later find a \emph{global} embedding for; and (ii) the learned gamma mapper.
We first discuss the forward rendering pass and then provide details about our training process. \Cref{fig:framework}
provides an overview of our method.

\subsection{Inference}

To render a scene, we use the conventional NeRF~\cite{nerf} paradigm with an Instant Neural Graphics Primitives (InstantNGP)~\cite{ingp} backbone.
Specifically, given a 3D position, $\mathbf{x}$, and a direction, $\mathbf{d}$, we train a neural network that maps them into color, $\mathbf{c}$, and density, $\mathbf{\sigma}$.
Then, for a given pixel with position $(x,y)$ with corresponding ray, $\mathbf{r(t)}$, in conventional NeRF, the color of this pixel, $\mathcal{C}$, is rendered by integrating along this ray between depths $d_s$ and $d_n$:
\begin{equation} \label{col_eq}
  \hat{\mathcal{C}}(\mathbf{r}) = \int_{d_s}^{d_n} \mathcal{T}(d)\sigma(\mathbf{r}(d))\mathbf{c}(\mathbf{r}(d), \mathbf{d}) dd 
  ,
\end{equation}
where $\mathcal{T}$ is the transmittance defined as
\begin{equation}
    \mathcal{T}(d) = \exp( -\int_{d_s}^d \sigma(\mathbf{r}(s))ds)
    .
\end{equation}
While this formulation works well in the ideal case with no blur
in our training images, and when sensors are well modelled under the chosen
sensor simplifications, it is insufficient when sensor modeling is critical, such as when blur is present.

\paragraph{Per-time embeddings}
To encompass the sensor modeling errors, we introduce a per-time embedding.
This strategy has previously been applied in other NeRF applications to model per-view illumination changes or to model transient objects~\cite{wildnerf}, and is also suggested in NeRFStudio~\cite{nerfstudio} to accommodate camera auto white balancing.
Here, we utilize it to model characteristics of the sensors at each capture time that are not included in the typical camera model.
To model how the world is perceived by the sensors
and not the underlying geometry, we pass the per-time embedding only to the color branch of the neural field; see \Cref{fig:framework}.
When rendering a frame at $t$, we add a $D$-dimensional learnable embedding, $\mathbf{E}_{t}\in \mathbb{R}^D$,
to \cref{col_eq}:
\begin{equation} \label{emb_eq}
    \hat{\mathcal{C}}(\mathbf{r}, t) = \int_{d_s}^{d_n} \mathcal{T}(d)\sigma(\mathbf{r}(d))\mathbf{c}(\mathbf{r}(d), \mathbf{d}, \mathbf{E}_{t}) dd
    .
\end{equation}
These learnable embeddings %
give the network the freedom to model the sensor imperfections.

\paragraph{Rendering events}
To obtain event data, we follow the definition of events.
That is, an event occurs if the logarithm of the intensity of a pixel, $\log(I)$, between a time interval defined by two times, $t_s$ and $t_e$, is greater than some threshold, $\omega$, at some pixel $x', y'$.
We thus write:
\begin{equation} \label{eq:event_model}
    |\log(I_{t_e}) - \log(I_{t_s})| \geq \omega
    .
\end{equation}
At instances when this condition is satisfied, in either direction, we obtain an event tuple 
\begin{equation} \label{eq:event_tuple}
    e = (x', y', t_e, p)
    ,
\end{equation}
where $p \in \{-1, 1\}$ is the sign (polarity) of the log intensity change.
As our focus is on  %
static NeRF applications, these intensity changes typically arise from camera motion.

However, a straightforward application of \cref{eq:event_model}, that is, setting $I = \rgbgray(\hat{\mathcal{C}})$, where $\rgbgray$ is the RGB to grayscale conversion function, assumes that the camera sensor response functions are the same for the RGB and event cameras.
To account for the sensor response differences, we use a gamma mapping: 
\begin{equation} \label{eq:gamma_model}
    I = \rgbgray(\hat{\mathcal{C}})^c
    ,   
\end{equation}
where $c$ is a learnable mapping parameter.
In contrast, others learn the response function with MLPs (e.g.,  EvDeblurNeRF~\cite{evDeblurNeRF})
but here, we opt for a simple gamma mapping, the traditional mapping function used to model sensor response functions.

\subsection{Training}

To train our network we rely on two losses: one for the RGB data and one for the event data.
Our total loss is given by
\begin{equation} \label{eq:total_loss}
    \mathcal{L} = \mathcal{L}_{rgb} + \lambda_{evs} \mathcal{L}_{evs}
    ,
\end{equation}
where $\mathcal{L}_{*}$ is the loss for the respective data, and $\lambda_{evs}$ is the hyperparameter that balances the two losses.
We next explain the individual losses.

\paragraph{RGB loss -- $\mathcal{L}_{rgb}$}
For the RGB image, we use the standard mean squared error loss, but with motion blur included.
Specifically, following previous work \cite{deblurnerf, BADNeRF}, we model the blur as a continuous camera pose, $\mathbf{P(t)}$.
Denoting the exposure as $\tau$, and the pathway a ray $\mathbf{r}$ takes as $\mathbf{r}(\mathbf{P(t)})$, we express the rendered blurry pixel color as:
\begin{equation}
    \hat{\mathcal{B}}(\mathbf{r}, t) = \int_\tau \hat{\mathcal{C}}(\mathbf{r}(\mathbf{P(t)}), t) dt
    .
\end{equation}
In practice, this integral is replaced with a Monte Carlo estimate:
\begin{equation} \label{eq:blur_model}
    \hat{B} = \sum_{i = 1}^n \hat{\mathcal{C}}(\mathbf{r}(\mathbf{P(t_i)}), t_i)
    ,
\end{equation}
where $t_i \sim U[t - \frac{\tau}{2}, t + \frac{\tau}{2}]$, $U$ is the uniform distribution and $n$ is the number of samples. 
We then use these blurry RGB reconstructions to supervise the network's blurry approximation:
\begin{equation}
    \mathcal{L}_{rgb} = \mathbb{E}_{\mathbf{r}, t}\left[\left\|\hat{\mathcal{B}}(\mathbf{r}, t) - \mathcal{B}_{gt}\right\|^2\right]
    ,
    \label{eq:rgb_loss_blurry}
\end{equation}
where $\mathcal{B}_{gt}$ is the ground truth blurry RGB image.

Note here how the supervision in \cref{eq:rgb_loss_blurry} already takes into account the motion blur, and thus allows the network to learn a sharp signal that corresponds to the blur.
This, however, is inherently an under-constrained problem, as there are multiple sharp NeRFs that can render the same blur image.
Thus, while deblurNeRF methods~\cite{deblurnerf, BADNeRF, dpnerf} provide a sharper reconstruction than NeRF, their reconstructions can still be blurry, as shown already in \Cref{fig:teaser}.

\paragraph{Event loss -- $\mathcal{L}_{evs}$}
To supervise events, we use brightness increment images (BII)~\cite{eklt}.
Specifically, for each pixel at position $(x, y)$ at time $t$, we sum the intensity changes that occur within a time window $\Delta t$:
\begin{equation} \label{eq:BII}
    I_{ev}[x, y, t] = \omega_0 \sum_{e_i \in \mathcal{E}} p_i
    ,  
\end{equation}
where considering the event tuples in \cref{eq:event_tuple}, we have:
\begin{equation}
    \quad \mathcal{E} = \{e_i| x = x'_i, y = y'_i, |t_{e_i} - t| \leq \Delta t \}
    ,
\end{equation}
$\omega_0 = 0.2$ is the default value used previously~\cite{esim, e2nerf}, and $\Delta t = 2.5$~ms is the optimal value found in prior work~\cite{eventDriving}.

With the brightness increment image, $I_{ev}[x, y, t]$, we supervise our network with the conventional event loss~\cite{enerf} that minimizes the difference between the events that our network renders and the ground truth events.
With the gamma mapped image, $I$, from \cref{eq:gamma_model}, we write:
\begin{equation}
    \begin{split}
    \mathcal{L}_{evs} = \mathbb{E}_{\mathbf{r},(t_s,t_e)}
    \Bigg[
    \bigg\|
        \log(I(\mathbf{r},t_e)) - 
        \log(I(\mathbf{r},t_s)) \\
        - I_{ev}\Big[x_{\mathbf{r}}, y_{\mathbf{r}}, \frac{t_e+t_s}{2}\Big]
        \bigg\|^2
        \Bigg],
    \end{split}
    \label{eq:event_loss}
\end{equation}
where $t_e$ and $t_s$ are the event times for each event stored in $I_{ev}[x, y, t]$, and $x_{\mathbf{r}}$ and $y_{\mathbf{r}}$ are the pixel coordinates corresponding to a ray $\mathbf{r}$.

\def \imgwidth {0.195\linewidth}

\begin{figure*}
    \centering
    \begin{subfigure}{\imgwidth}
        \includegraphics[width=\linewidth]{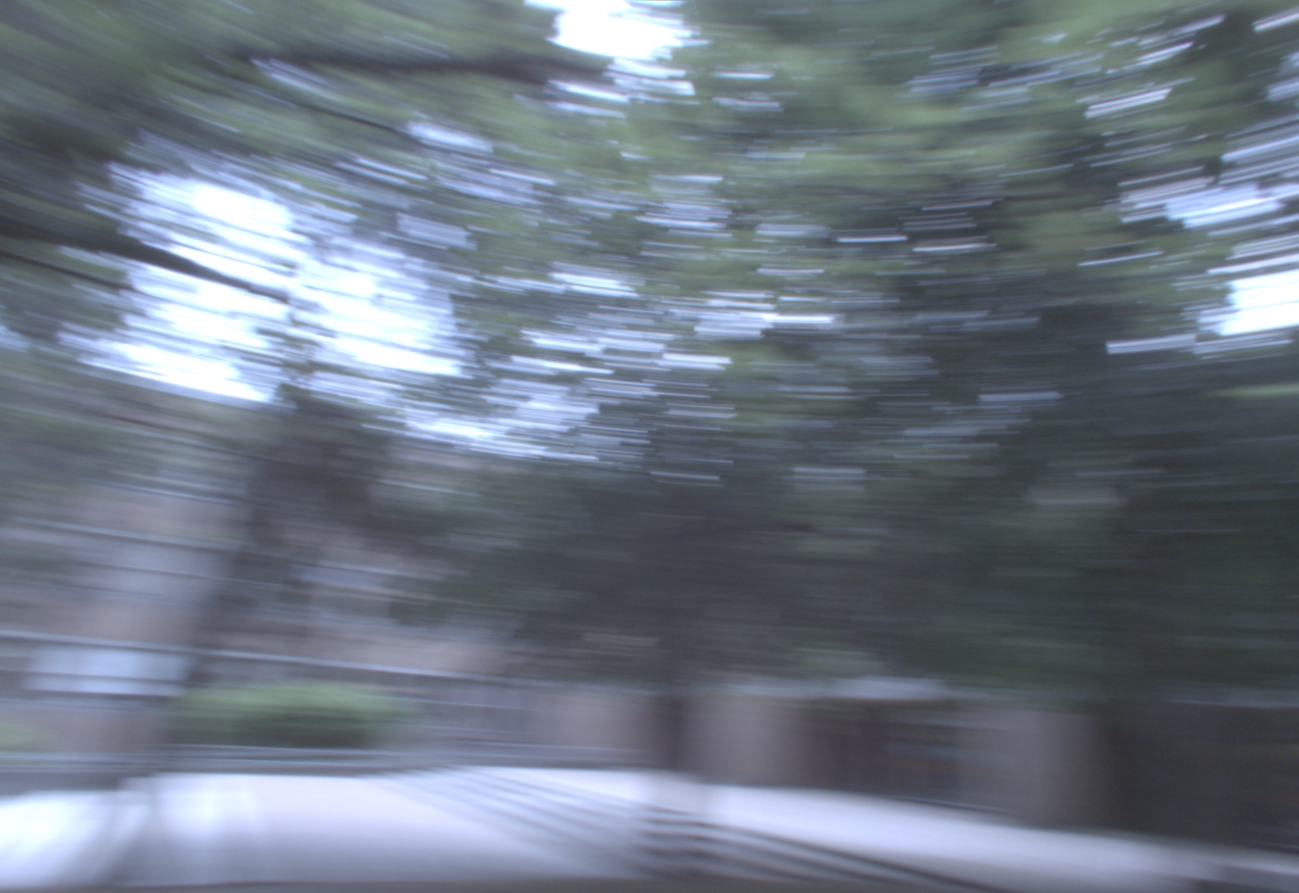}
    \end{subfigure}
    \begin{subfigure}{\imgwidth}
        \includegraphics[width=\linewidth]{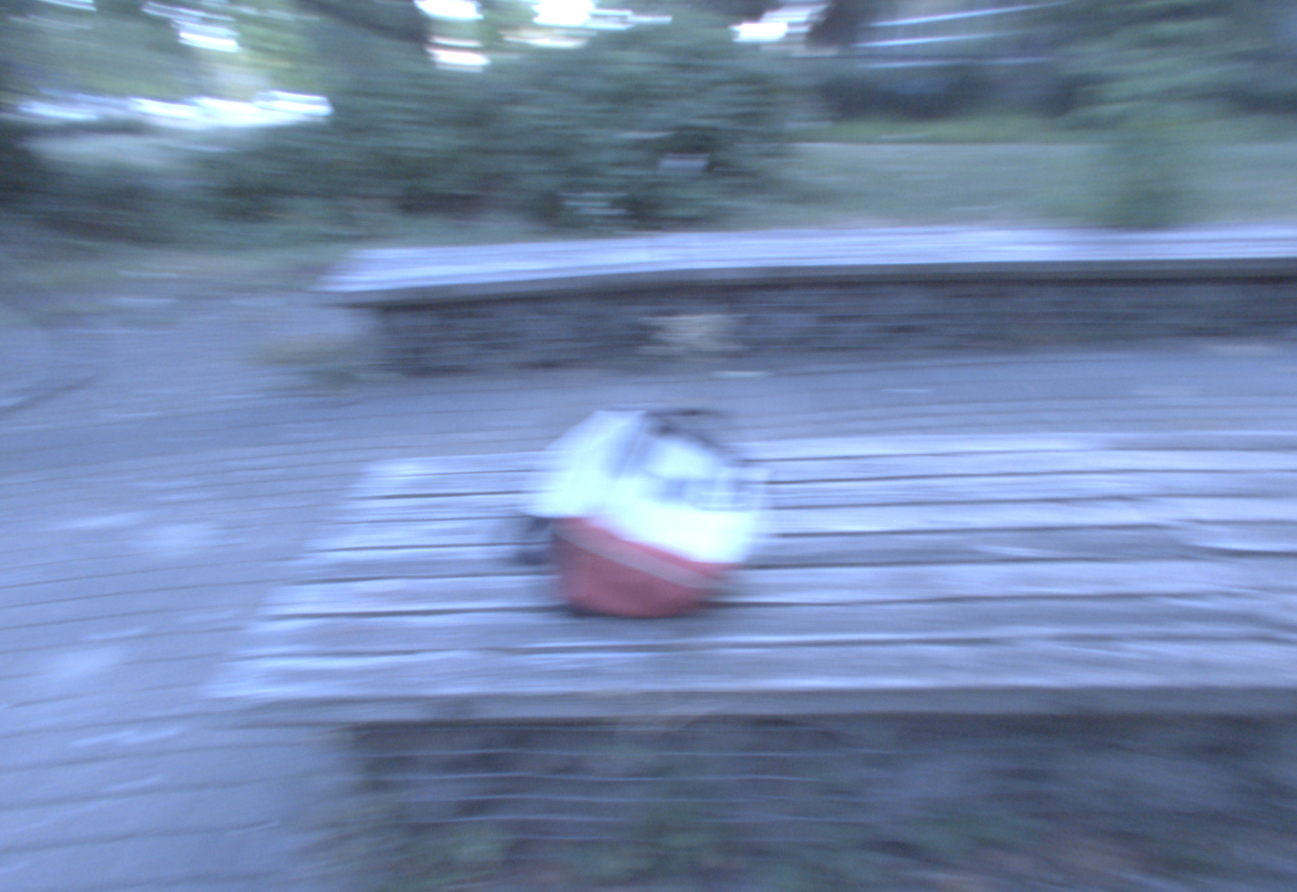}
    \end{subfigure}
    \begin{subfigure}{\imgwidth}
        \includegraphics[width=\linewidth]{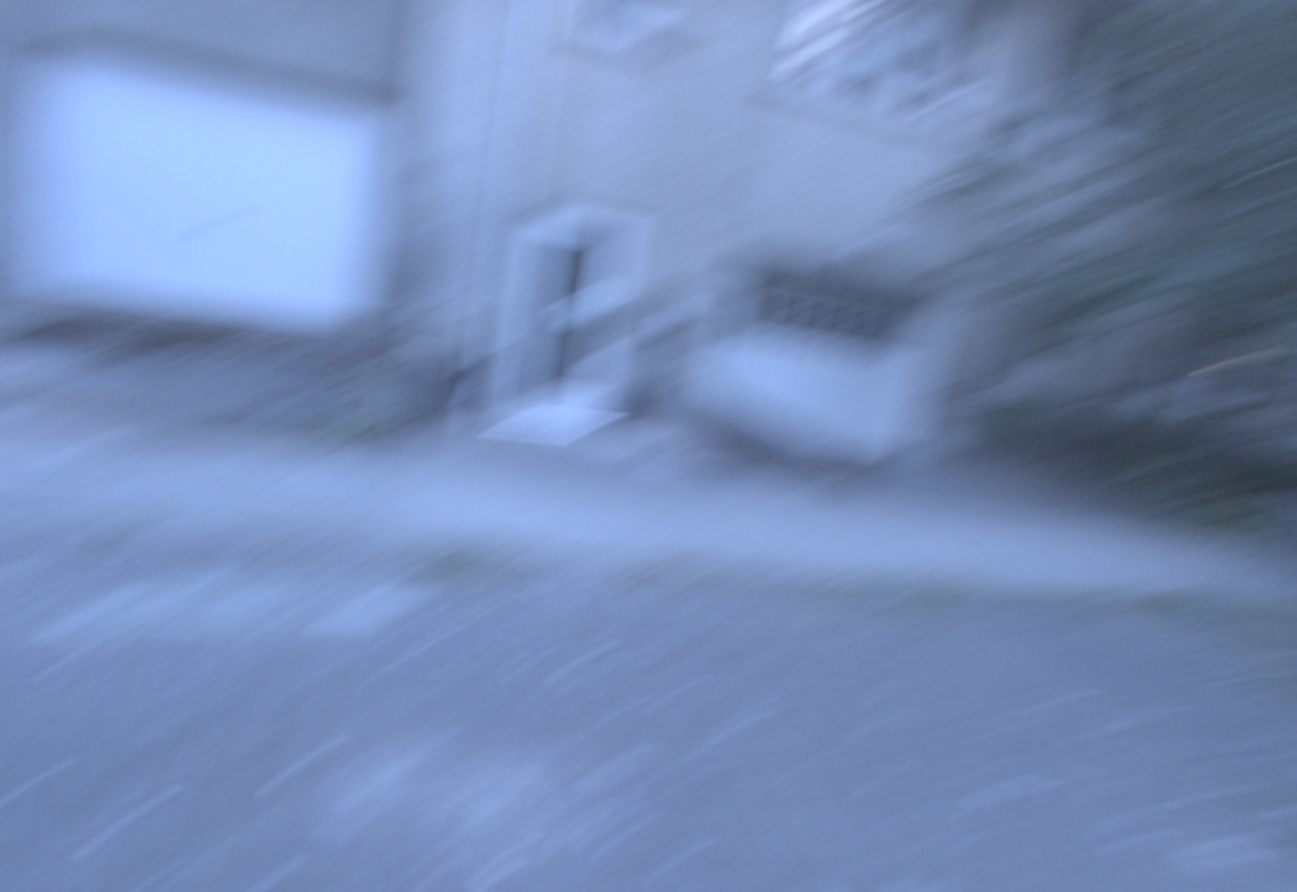}
    \end{subfigure}
    \begin{subfigure}{\imgwidth}
        \includegraphics[width=\linewidth]{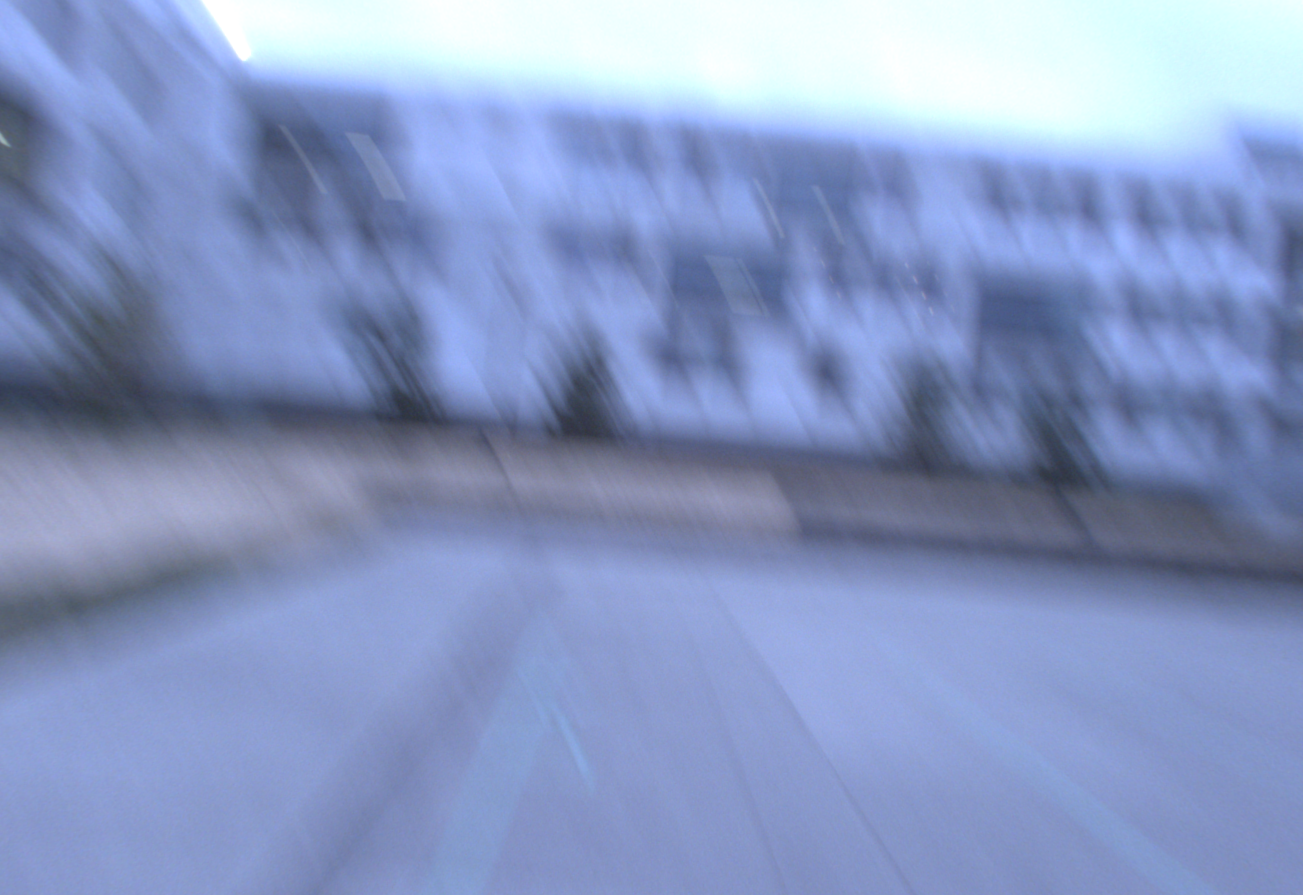}
    \end{subfigure}
    \begin{subfigure}{\imgwidth}
        \includegraphics[width=\linewidth]{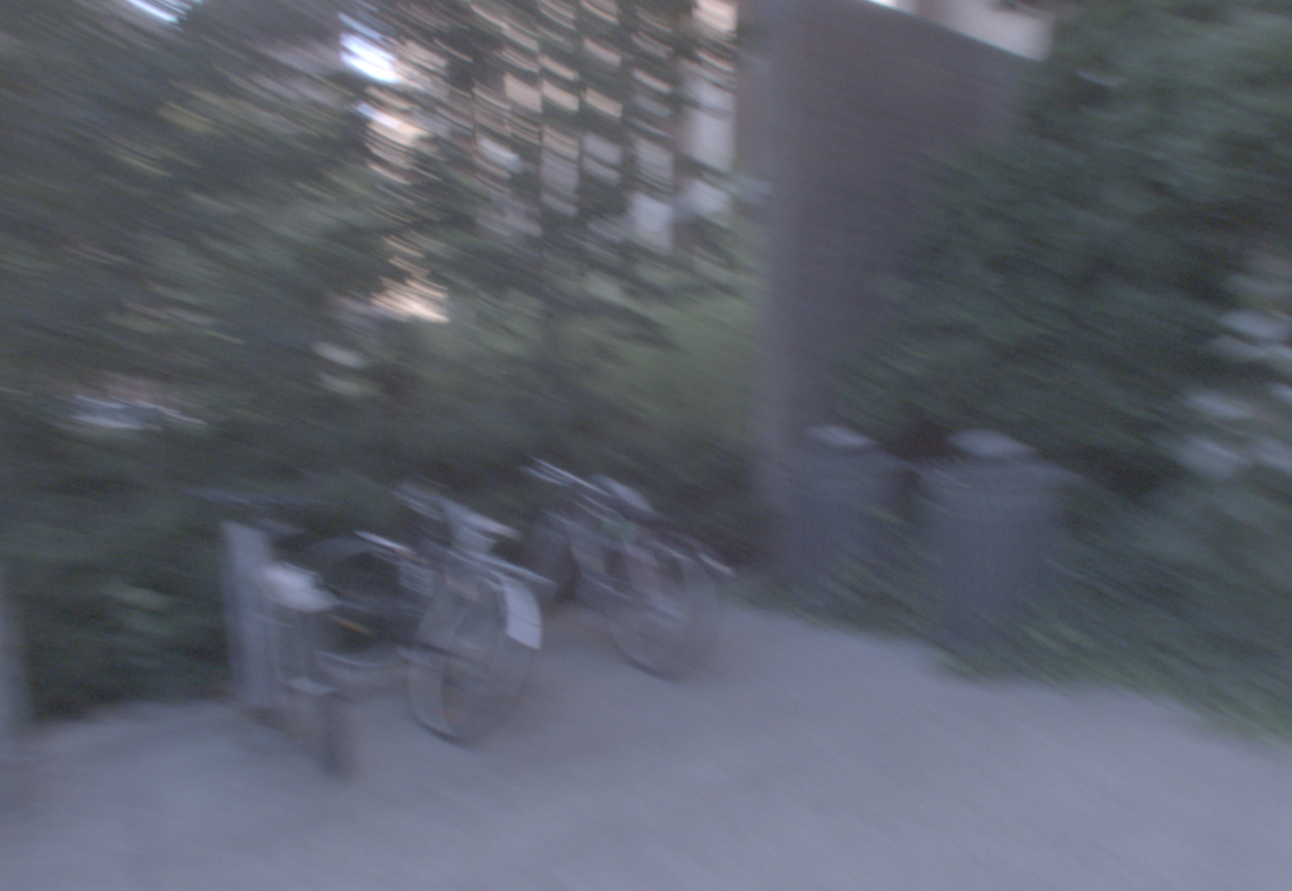}
    \end{subfigure} \\
    \begin{subfigure}{\imgwidth}
        \includegraphics[width=\linewidth]{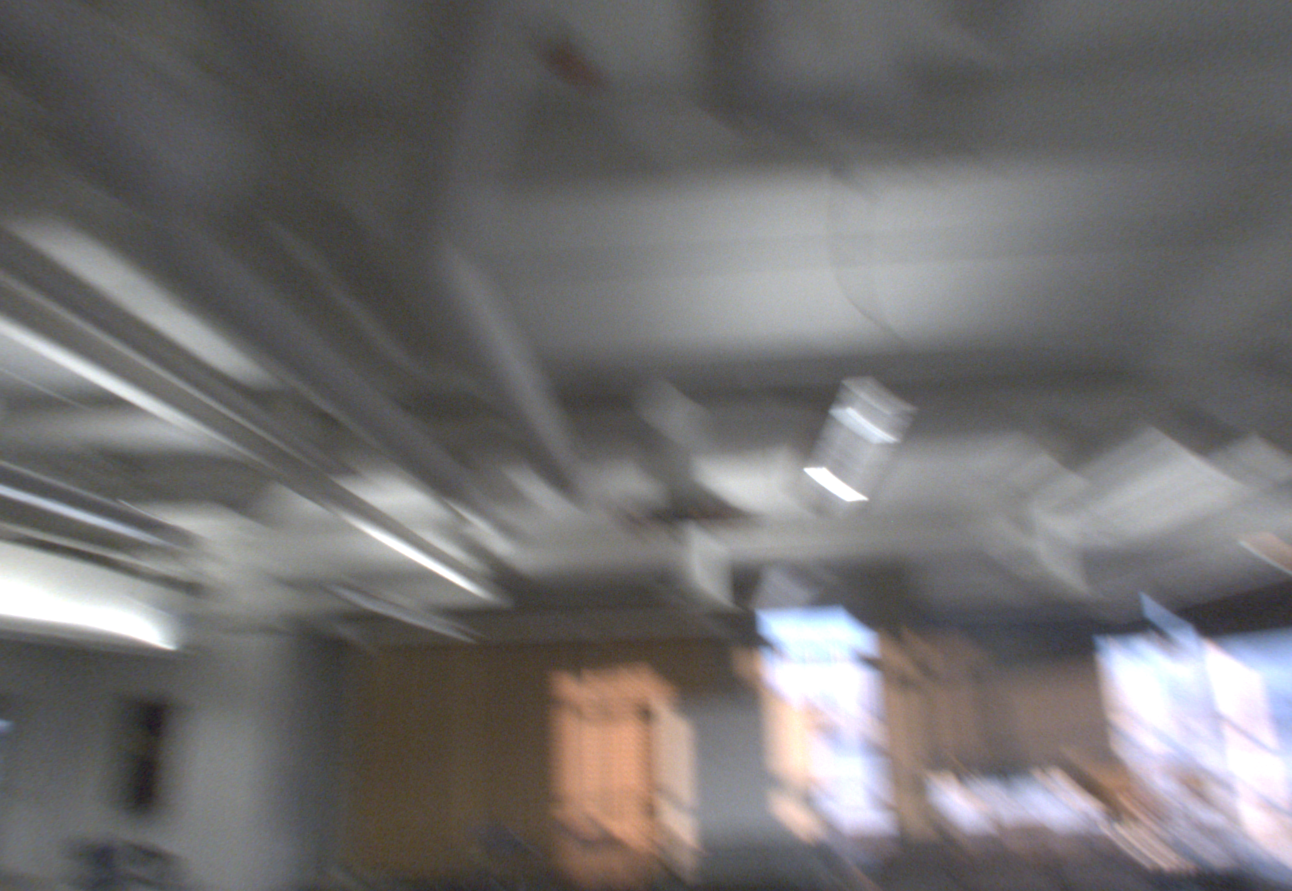}
    \end{subfigure}
    \begin{subfigure}{\imgwidth}
        \includegraphics[width=\linewidth]{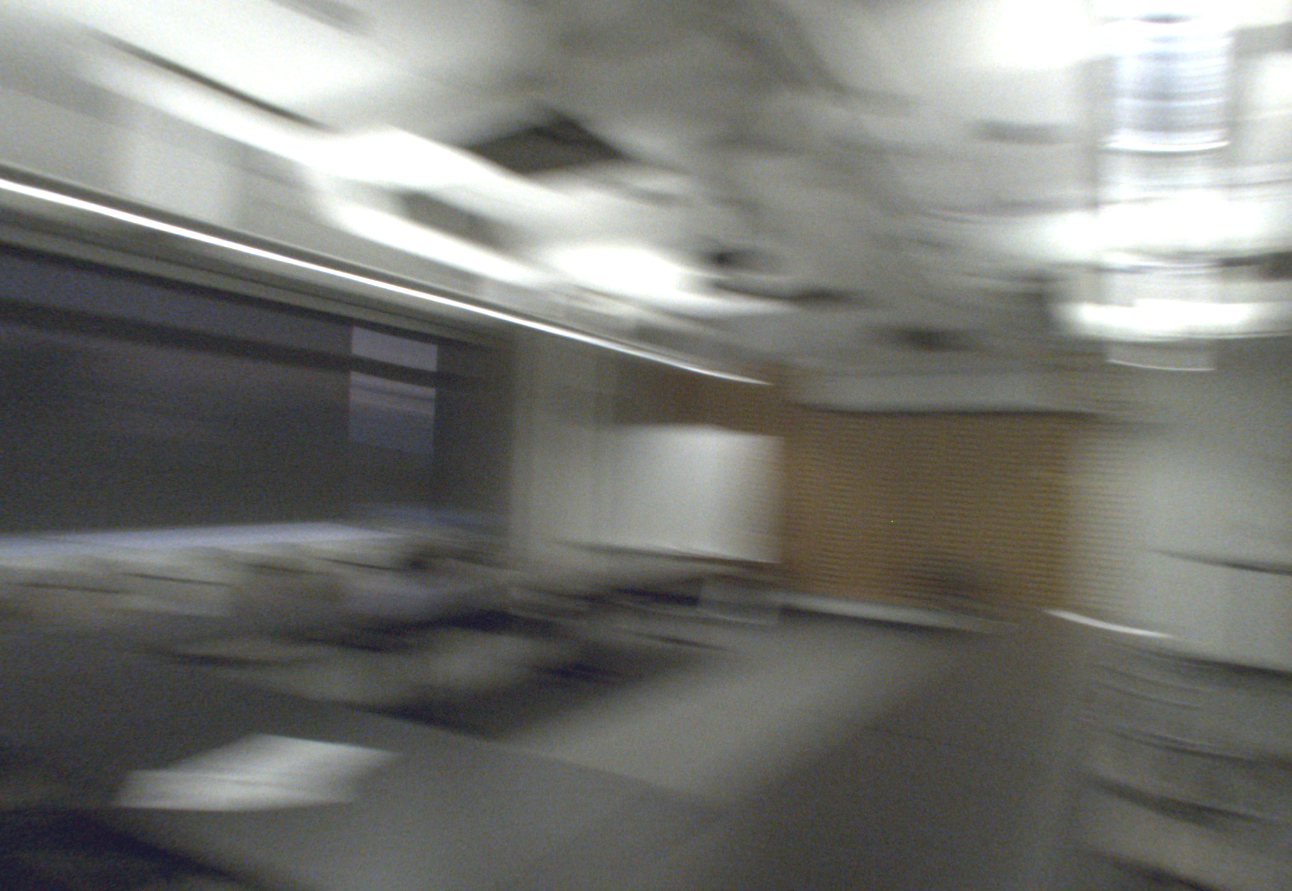}
    \end{subfigure}
    \begin{subfigure}{\imgwidth}
        \includegraphics[width=\linewidth]{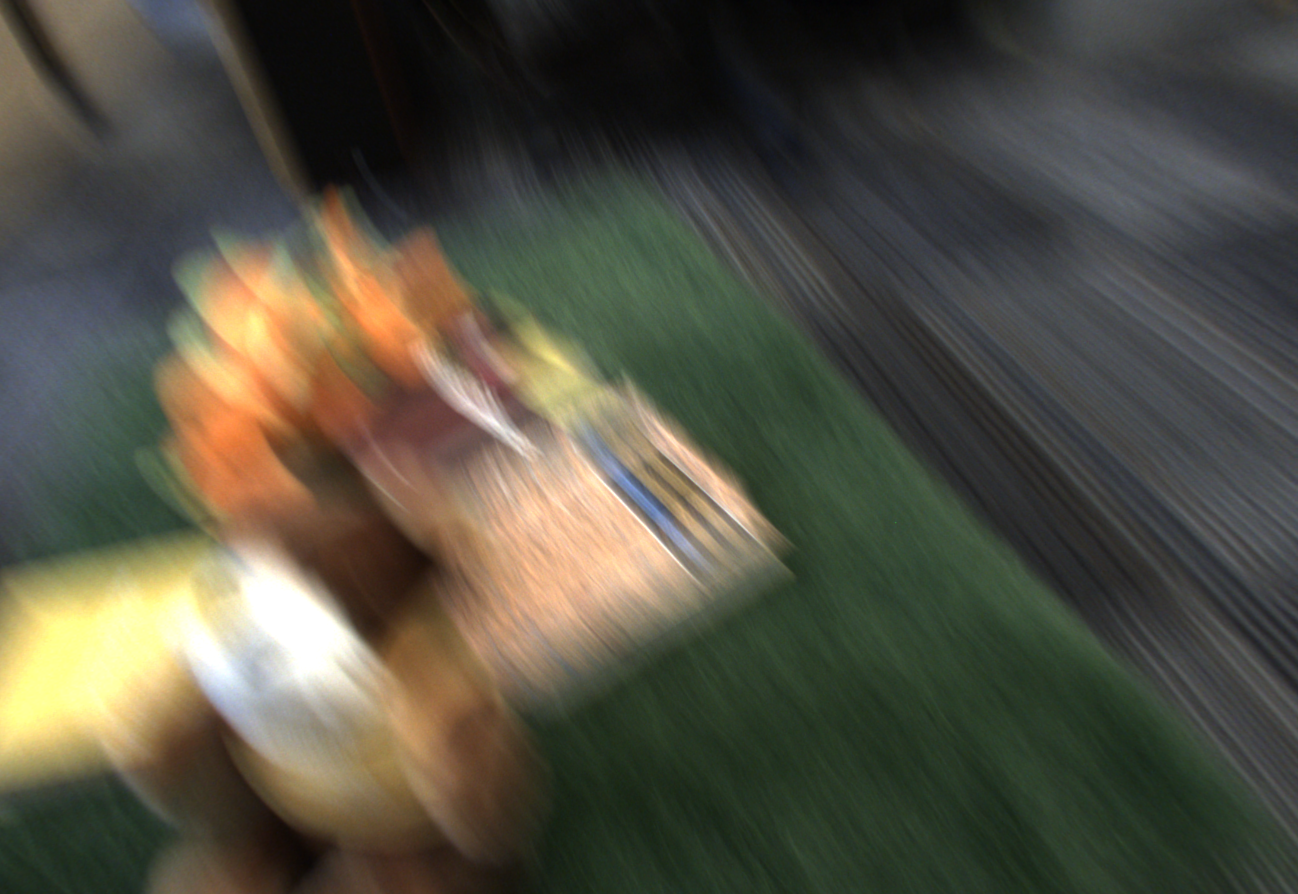}
    \end{subfigure}
    \begin{subfigure}{\imgwidth}
        \includegraphics[width=\linewidth]{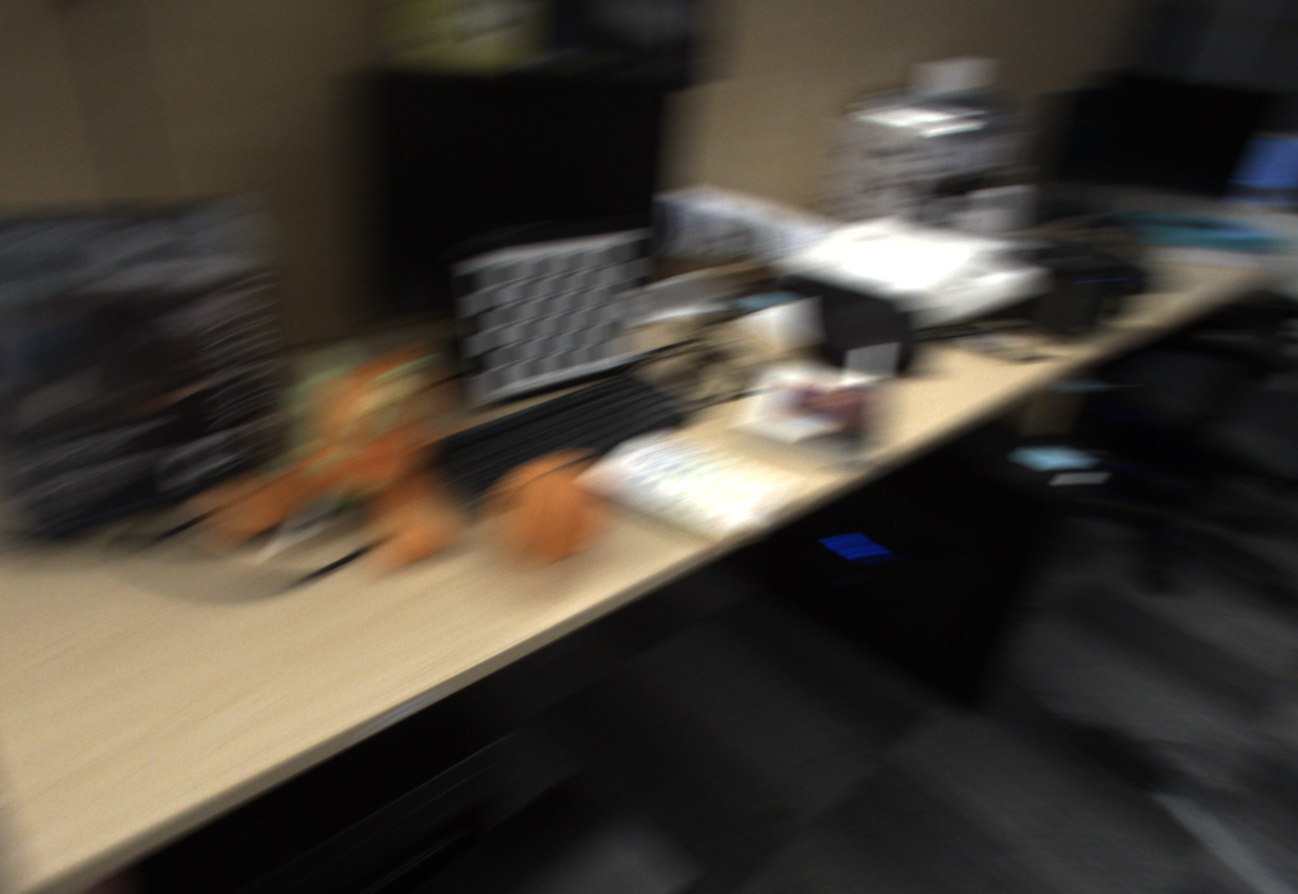}
    \end{subfigure}
    \begin{subfigure}{\imgwidth}
        \includegraphics[width=\linewidth]{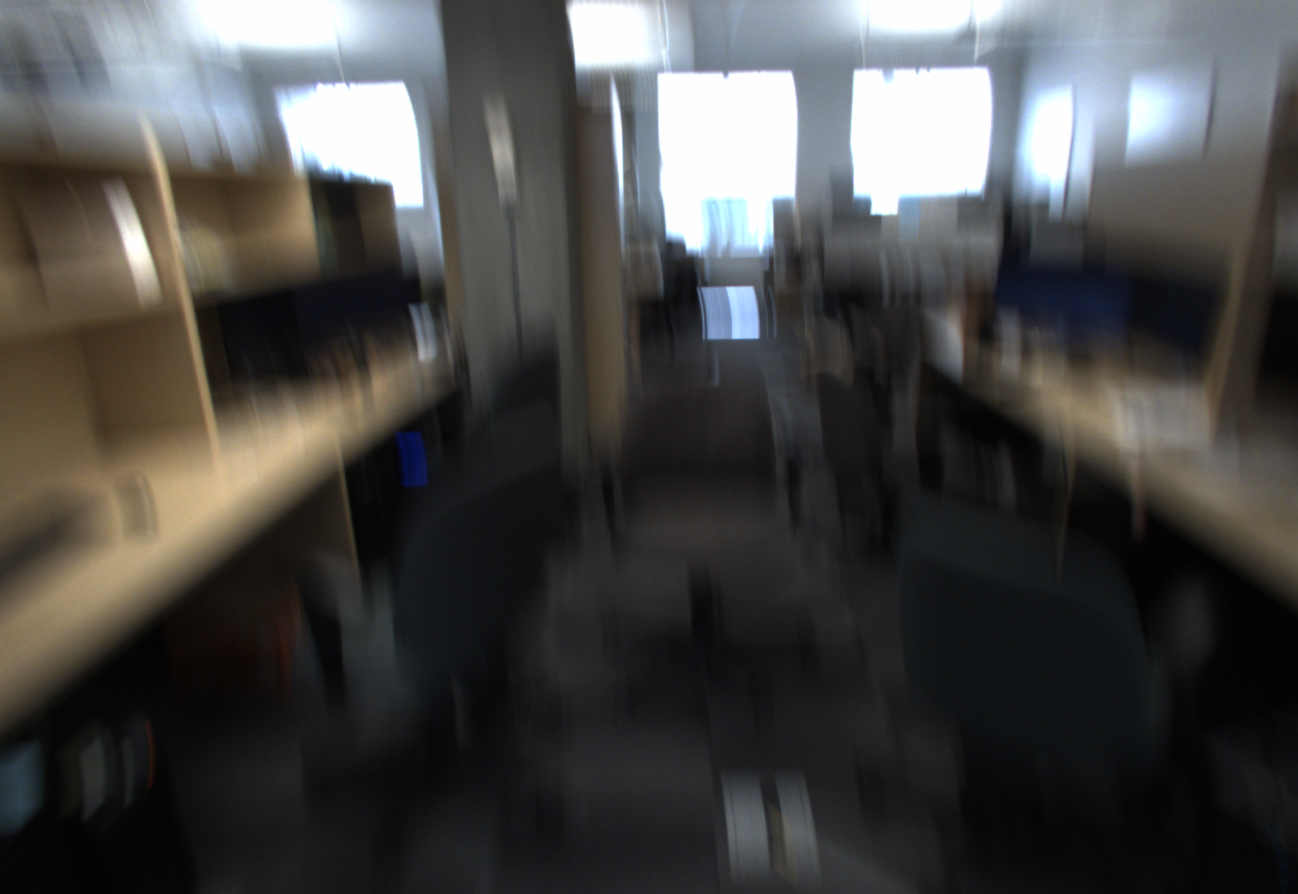}
    \end{subfigure}
    \caption{{\bf Sample RGB frames from each scene in our dataset -- }
    Our dataset consists of {\bf (top row)} five outdoor scenes and {\bf (bottom row)} five indoor scenes.  The substantial image blur is caused by rapid camera movements.  
    }
\end{figure*}

\begin{figure}
    \centering
    \includegraphics[width=0.5\linewidth, trim=0 50 0 80, clip]{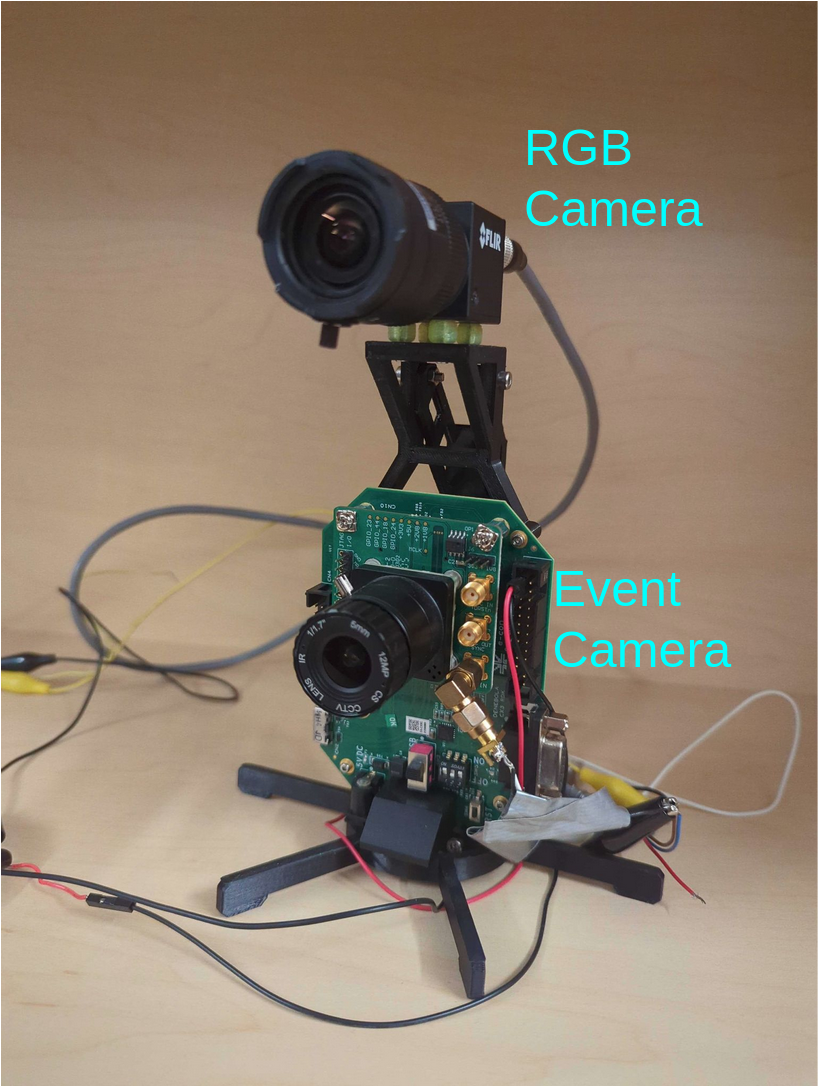}
    \includegraphics[width=0.455\linewidth, trim=0 50 0 80, clip]{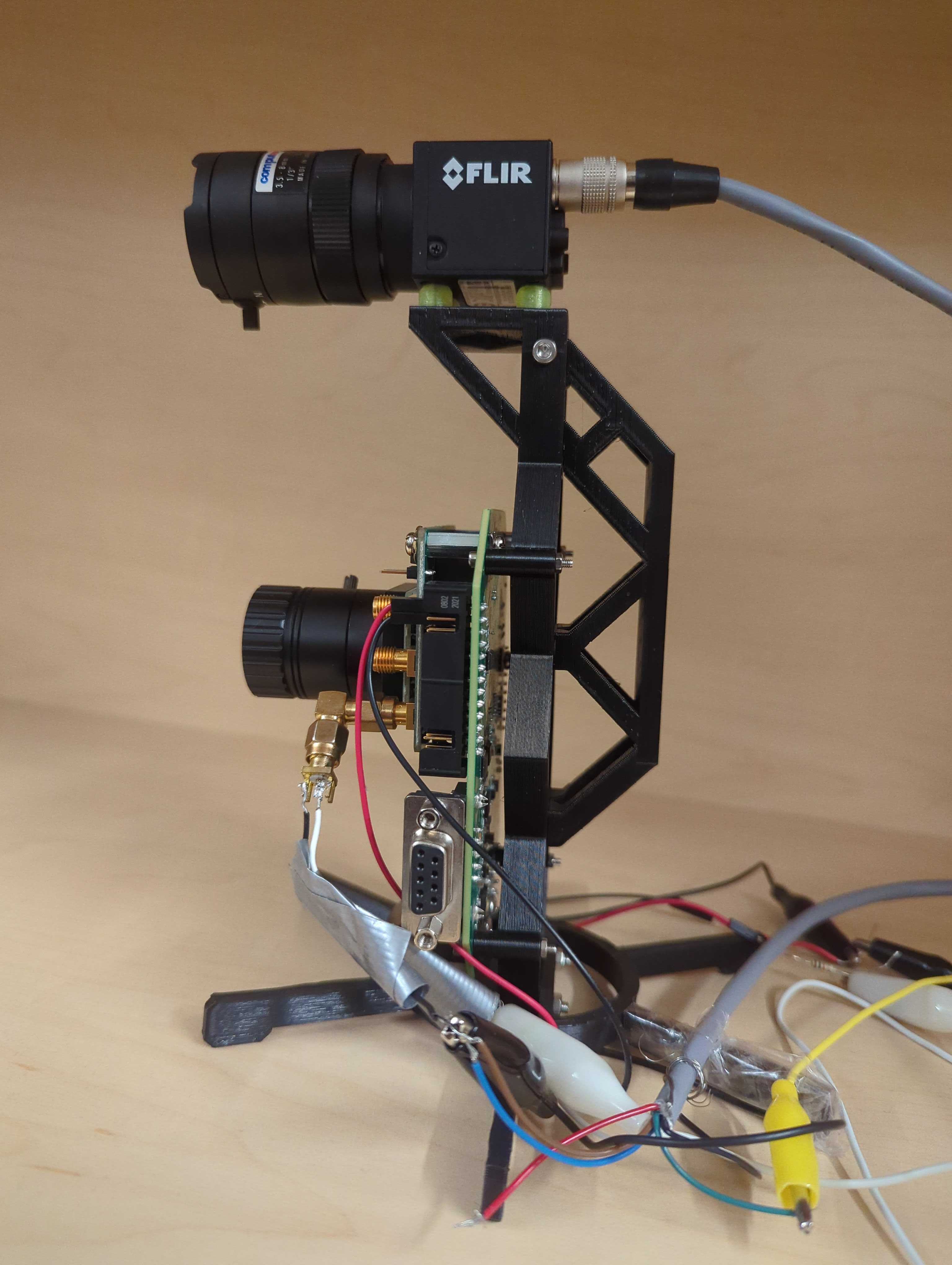}
    \caption{{\bf Capture rig -- }
    We 3D print a stereo casing that holds a GigE Blackfly S camera and a Prophesee EVK-3 HD camera.
    }
    \label{fig:rig}
\end{figure}

\paragraph{Finding the \emph{global} embedding}
\label{sec:global_embedding}
Finally, once the network is trained, we note that to render a novel view, we do not have the per-time embedding for a new frame.
We thus find a \emph{global} embedding, $\mathbf{E}_* \in \mathbb{R}^D$, that works well for all frames.
To achieve this, we freeze the entire network except the global embedding weights and then retrain for a few thousand steps to determine the global embedding for evaluation.
In doing so, we utilize only the RGB loss, $\mathcal{L}_{rgb}$, to find the global embedding, as we empirically found that using the event loss, $\mathcal{L}_{evs}$, did not work well---we ablate this in \Cref{sec:ablation}.

\paragraph{Camera optimization}
\label{sec:cam_opt}
Camera poses recovered from blurry images are often inaccurate. 
To mitigate this, we allow
both the RGB and the event camera poses to be optimized. 
For both cameras, we parameterize the camera pose with exponential maps and optimize the camera poses directly with gradients from the respective losses.
\begin{itemize}
    \item {\bf RGB cameras:} 
    As the RGB loss utilizes blurry images, it is important to consider `intermediate' cameras for which we do not have ground-truth images for---as in BADNeRF~\cite{BADNeRF}, we assume a linear trajectory for the RGB camera using the exponential map parameterization.
    We then optimize the poses of the cameras that we have ground-truth images for directly with gradients from the RGB loss. 
    \item {\bf Event cameras:} For the event camera positions we discretize the trajectory into individual cameras for each brightness increment image. 
    We initially set the event camera pose as linear interpolations of the RGB camera pose. 
    We then optimize them according to the event loss, again assuming a linear trajectory between cameras.

\end{itemize}
As the initial reconstructions are unstable, we only enable camera optimization after the first 10K steps of training.

\def \qualwidth {0.155}

\begin{figure*}
    \centering
    \setlength{\tabcolsep}{2pt}
    \footnotesize
    \begin{tabular}{c c c c c c c}
        & Ground Truth & \makecell{BADNeRF~\cite{BADNeRF}\\(RGB Only)} & \makecell{BADNeRF~\cite{BADNeRF} + Emb.\\(RGB Only)} & \makecell{E2NeRF~\cite{e2nerf}\\(RGB + Events)}& \makecell{Our Method w/o Emb\\(RGB + Events)} & \makecell{{\bf Our Method}\\(RGB + Events)} \\

        \rotatebox{90}{\hspace{3em}House} &
        {\includegraphics[width=\qualwidth\linewidth]{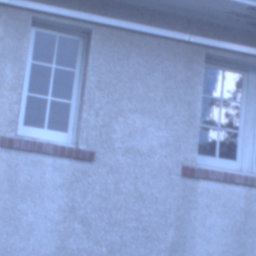}} &
        {\includegraphics[width=\qualwidth\linewidth]{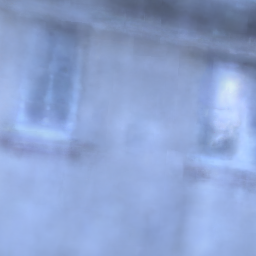}} &
        {\includegraphics[width=\qualwidth\linewidth]{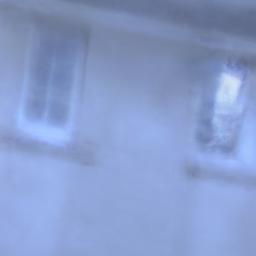}} & 
        {\includegraphics[width=\qualwidth\linewidth]{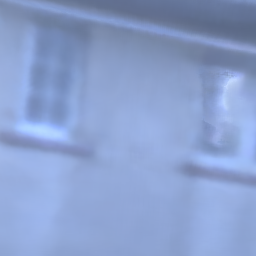}} & 
        {\includegraphics[width=\qualwidth\linewidth]{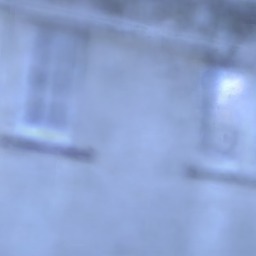}} &
        {\includegraphics[width=\qualwidth\linewidth]{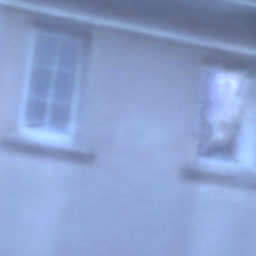}} \\

        \rotatebox{90}{\hspace{2em}Courtyard} &
        {\includegraphics[width=\qualwidth\linewidth]{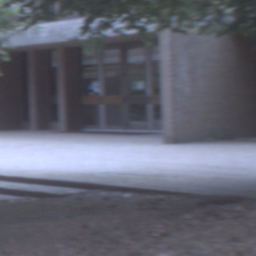}} &
        {\includegraphics[width=\qualwidth\linewidth]{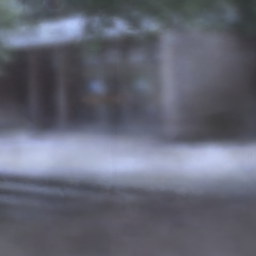}} &
        {\includegraphics[width=\qualwidth\linewidth]{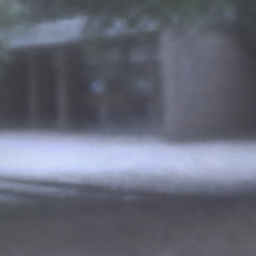}} &
        {\includegraphics[width=\qualwidth\linewidth]{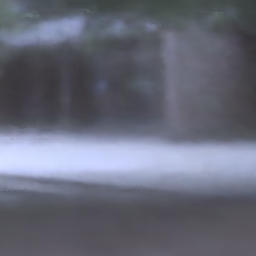}} &
        {\includegraphics[width=\qualwidth\linewidth]{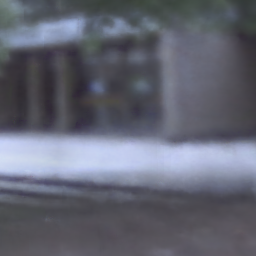}} &
        {\includegraphics[width=\qualwidth\linewidth]{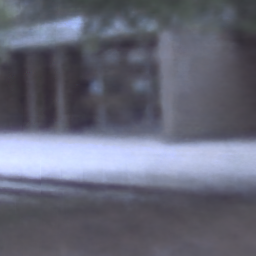}} \\

        \rotatebox{90}{\hspace{2em}Dragon\_Max} &
        {\includegraphics[width=\qualwidth\linewidth]{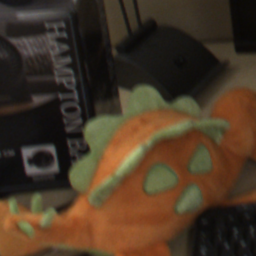}} &
        {\includegraphics[width=\qualwidth\linewidth]{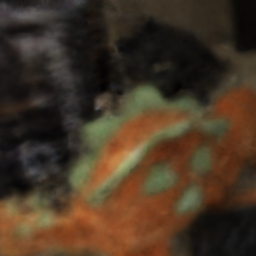}} &
        {\includegraphics[width=\qualwidth\linewidth]{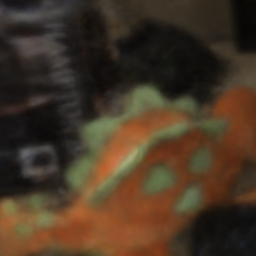}} & 
        {\includegraphics[width=\qualwidth\linewidth]{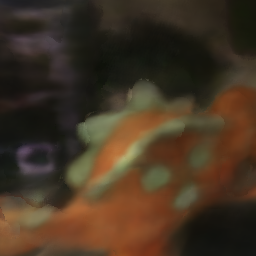}} & 
        {\includegraphics[width=\qualwidth\linewidth]{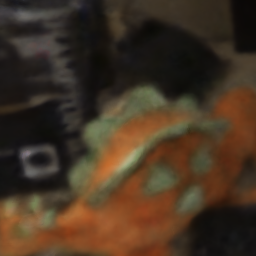}} &
        {\includegraphics[width=\qualwidth\linewidth]{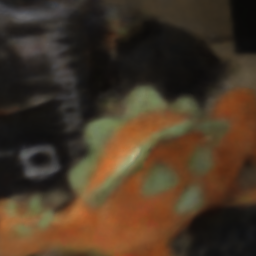}} \\

        \rotatebox{90}{\hspace{1em}Presentation\_Room} &
        {\includegraphics[width=\qualwidth\linewidth]{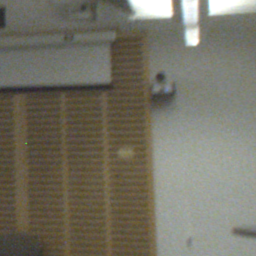}} &
        {\includegraphics[width=\qualwidth\linewidth]{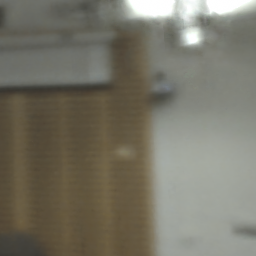}} &
        {\includegraphics[width=\qualwidth\linewidth]{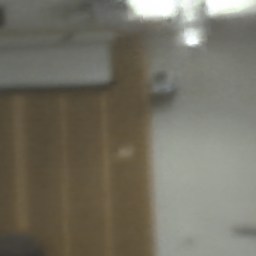}} &
        {\includegraphics[width=\qualwidth\linewidth]{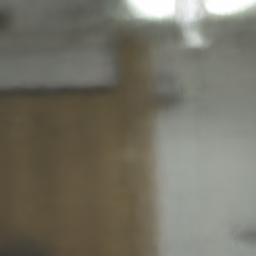}} &
        {\includegraphics[width=\qualwidth\linewidth]{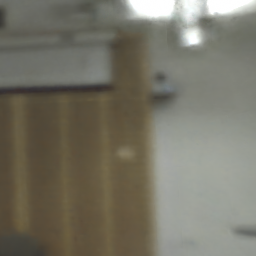}} &
        {\includegraphics[width=\qualwidth\linewidth]{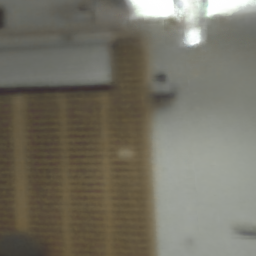}} \\

    \end{tabular}
    \caption{\textbf{Qualitative examples -- } 
    We show qualitative examples of zoomed-in reconstruction cutouts.
    Our method provides the sharpest reconstructions.
    Interestingly, BADNeRF~\cite{BADNeRF}, combined with our embedding strategy, also provides clear results, being on par or slightly better than even when event data is used.
    The best results, however, are obtained with our method
    where, RGB images are used together with event data.
    }
    \label{fig:qualitative}
\end{figure*}

\section{Data Collection}

To validate our method, we collect our own dataset containing five indoor and five outdoor scenes. 

\paragraph{Capture rig}
While various other datasets exist, they are either synthetic, or contain data collected from variants of the DAVIS~346 camera that captures aligned RGB and events. 
The assumption of aligned streams is a strong one, limited to certain devices, one that our method relaxes.
Furthermore, typically, these devices operate on a lower spatial resolution to support both modalities.
Hence, we opt for a more versatile capture setup, where, as shown in \Cref{fig:rig}, we 3D print a stereo casing and use a GigE Blackfly S \cite{flir_blackfly_s_2024} camera to capture RGB images and a Prophesee EVK-3 HD camera \cite{evk3-hd} to capture events.
Each device provides high-resolution data, with the RGB camera capturing $1440 \times 1080$ images and the event camera capturing $1280 \times 720$ events. 
The two devices are temporarily hardware-synced by triggers sent from the RGB camera.

\paragraph{Capture protocol}
We capture our dataset by hand-holding the stereo rig in various locations.
Each scene is approximately 20 seconds long, with the first 13 seconds being blurry with fast swinging motions and the last seven seconds exhibiting slow stable camera movement to provide clear frames for evaluation.

\paragraph{Calibration}
To calibrate the two cameras (i.e., recover the intrinsics and relative extrinsics), we use the standard checkerboard calibration pattern and slowly move our stereo rig to create events around the edges of the checkerboard.

We use the model from E2Calib~\cite{e2calib} to map event streams to event images and then stereo calibration from OpenCV~\cite{opencv} to recover the intrinsics and extrinsics.

\paragraph{Camera pose}
To recover the camera poses, we use COLMAP~\cite{colmap1, colmap2, colmap3}.
To allow COLMAP to work well with blurry images, we modify its default parameters and loosen the constraints on initialization and match filtering.
Specifically, we reduce the initialization constraint to have 28 inliers with a minimum of five degrees. 
We also reduce the observation filter constraint by setting the reprojection error to 12 pixels and reducing the triangulation filter's minimum angle to 0.08 radian.

Furthermore, as the extrinsics between the two cameras for our calibrated stereo rig are provided in their own scene scale (specifically metric scale), we need to also recover the scene scale.
To do this, we manually triangulate three points from the RGB images and annotate the corresponding keypoints in the event brightness increment images (BII).

We then use Powell's method~\cite{powell} to minimize the reprojection error of the triangulated points to the corresponding keypoints in the brightness increment images, which gives us the scene scale.
Finally, we obtain the event camera poses by applying the relative extrinsics to the RGB camera poses, with the scene scale.

\section{Results}
\label{sec:results}

\begin{table*}
\centering
\setlength{\tabcolsep}{2pt}
\resizebox{\textwidth}{!}{%
\begin{tabular}{@{}l ccc ccc ccc ccc ccc cccc@{}}  %
\toprule
& \multicolumn{3}{c}{Courtyard}  
& \multicolumn{3}{c}{Bag} 
& \multicolumn{3}{c}{House}  
& \multicolumn{3}{c}{Engineer Building} 
& \multicolumn{3}{c}{Bicycle} 
& & \multicolumn{3}{c}{Average} \\ 
\cmidrule(r){2-4} 
\cmidrule(r){5-7}
\cmidrule(r){8-10}
\cmidrule(r){11-13}
\cmidrule(r){14-16}
\cmidrule{18-20}
& SSIM$\uparrow$ & PSNR$\uparrow$ & LPIPS$\downarrow$
& SSIM$\uparrow$ & PSNR$\uparrow$ & LPIPS$\downarrow$
& SSIM$\uparrow$ & PSNR$\uparrow$ & LPIPS$\downarrow$
& SSIM$\uparrow$ & PSNR$\uparrow$ & LPIPS$\downarrow$
& SSIM$\uparrow$ & PSNR$\uparrow$ & LPIPS$\downarrow$
& & SSIM$\uparrow$ & PSNR$\uparrow$ & LPIPS$\downarrow$ \\
\midrule
BADNeRF~\cite{BADNeRF}
& 0.712 & 21.618 & 0.475
& 0.735 & 26.450 & 0.380 
& 0.734 & 22.670 & 0.457
& 0.806 & 24.169 & 0.372 
& 0.759 & 25.775 & 0.398 
& & 0.749 & 24.137 & 0.416 \\
BADNeRF~\cite{BADNeRF} + Emb.                  
& 0.716 & \second{22.359} & \second{0.451} 
& \second{0.754} &  \second{26.895} & \best{0.357} 
& 0.751 & \second{24.227} & \second{0.420} 
& \second{0.823} & \second{25.446} & \second{0.339} 
& 0.760 & \second{26.363} & 0.398 
& & \second{0.761} & \second{25.058} & \second{0.393} \\
\midrule
E2NeRF~\cite{e2nerf} 
& 0.691 & 21.904 & 0.592 
& 0.696 & 26.172 & 0.508
& 0.757 & 23.754 & 0.501
& 0.785 & 24.37 & 0.451
& 0.726 & 24.637 & 0.574 
& & 0.731 & 24.168 & 0.525\\
Our method w/o Emb.
& \second{0.719} & 22.314 & 0.472 
& 0.749 & 27.118 & 0.374 
& \second{0.758} & 23.855 & 0.442
& 0.819 & 25.179 &  0.346 
& \second{0.760} & 26.278 &  \second{0.390} 
& & 0.761 & 24.949 & 0.405 \\
Our method
& \best{0.723} & \best{22.660} & \best{0.442} 
& \best{0.760} & \best{27.575} & \second{0.367}
& \best{0.778} & \best{25.047} & \best{0.416} 
& \best{0.827} & \best{25.461} & \best{0.324} 
& \best{0.778} & \best{26.620} & \best{0.388} 
& & \best{0.772} & \best{25.473} & \best{0.388} \\
\toprule
& \multicolumn{3}{c}{Grad Lounge}  
& \multicolumn{3}{c}{Presentation Room} 
& \multicolumn{3}{c}{Teddy Grass}  
& \multicolumn{3}{c}{Dragon Max} 
& \multicolumn{3}{c}{Lab} 
& & \multicolumn{3}{c}{Average} \\ 
\cmidrule(r){2-4} 
\cmidrule(r){5-7}
\cmidrule(r){8-10}
\cmidrule{11-13}
\cmidrule{14-16}
\cmidrule{18-20}
& SSIM$\uparrow$ & PSNR$\uparrow$ & LPIPS$\downarrow$
& SSIM$\uparrow$ & PSNR$\uparrow$ & LPIPS$\downarrow$
& SSIM$\uparrow$ & PSNR$\uparrow$ & LPIPS$\downarrow$
& SSIM$\uparrow$ & PSNR$\uparrow$ & LPIPS$\downarrow$
& SSIM$\uparrow$ & PSNR$\uparrow$ & LPIPS$\downarrow$
& & SSIM$\uparrow$ & PSNR$\uparrow$ & LPIPS$\downarrow$ \\
\midrule
BADNeRF~\cite{BADNeRF}
& 0.737 & 24.918 & 0.517 
& 0.692 & 24.706 & 0.503
& 0.673 & 23.485 & 0.572 
& 0.826 & 25.330 & 0.373 
& 0.816 & 22.389 & 0.356
& & 0.749 & 24.166 & 0.464 \\
BADNeRF~\cite{BADNeRF} + Emb.
& \best{0.775} & 26.261 & \second{0.440}
& 0.692 & \second{25.623} & \second{0.475}
& \second{0.699} & \second{25.120} & \best{0.552} 
& 0.850 & \second{26.195} & 0.328 
& \second{0.841} & 22.970  & 0.319 
& & \second{0.771} & \second{25.250} & \second{0.423} \\
\midrule
E2NeRF~\cite{e2nerf} 
& 0.707 & 24.927 & 0.564
& 0.634 & 23.24 & 0.613
& 0.61 & 23.363 & 0.681
& 0.777 & 23.921 & 0.427
& 0.746 & 21.667 & 0.487
& & 0.695 & 23.424 & 0.554 \\
Our method w/o Emb.
& 0.760 & \second{26.506} & 0.471 
& \second{0.699} & 25.338 & 0.510 
& 0.692 & 25.128 & 0.601 
& \second{0.850} & 26.123 & \second{0.318} 
& 0.841 & \second{22.986} & \second{0.310}
& & 0.768 & 25.216 & 0.442 \\
Our method
& \second{0.773} & \best{26.589} & \best{0.434} 
& \best{0.610} & \best{25.845} & \best{0.465} 
& \best{0.701} & \best{25.618} & \second{0.570}
& \best{0.856} & \best{26.349} & \best{0.307} 
& \best{0.855} & \best{23.144} & \best{0.288} 
& & \best{0.777} & \best{25.509} & \best{0.413} \\
\bottomrule
\end{tabular}
    }
\caption{
{\bf Quantitative results for our dataset -- }
We report the quantitative metrics for our method and the baselines on the test views.
We report results both for \textbf{(top rows)} outdoor scenes and \textbf{(bottom rows)} indoor scenes.
We mark \best{best} and \second{second-best} results in each column.
Our method performs best in terms of all metrics. 
Interestingly, BADNeRF with our embedding strategy performs second-best, although it is trained only on RGB images.
This demonstrates using per-time embeddings to measure sensor modeling imperfections helps significantly.
In \cref{sec:ablation},
we further show how one obtains the global embedding for inference is also critical.
}
\label{tab:quantitative} 
\end{table*}

\subsection{Experimental setup}

\paragraph{Dataset}
In addition to our dataset, we further utilize the \evimo dataset for evaluation.
\evimo is originally designed for object segmentation, optical flow and Structure-from-Motion (SFM) tasks. 
As it is not a dataset designed for deblurring tasks, we manually examine all images in the sequences chosen for our experiments and manually select clear images to be held out for testing.
We choose three sequences from the dataset that have a sufficient number of clear images: \evimosceneA, \evimosceneB, \evimosceneC.
Because the dataset is not designed to test debluring, we note that the selected clear images may still contain slight blur.\footnote{To ensure the reproducibility of our results, we will release the data processing code for the selected scenes.}

\paragraph{Baselines}
We compare our method with the following:
\begin{itemize}
\item {\bf BADNeRF~\cite{BADNeRF}: } We reimplement their method in our framework but without the exposure time optimization, as in our data we have the exact exposure time.
We use this baseline to compare against the case when we do not use the event data.
\item {\bf E2NeRF~\cite{e2nerf}: } We use the official code to demonstrate the performance of a recent event-based NeRF method.
\item {\bf Ablations:} We further compare our method against ablations, such as without the embedding, without the mapper, and without the camera optimization to demonstrate the effectiveness of each component.
\end{itemize}

\paragraph{Metrics}
We use the following standard image quality evaluation metrics: %
Peak Signal-to-Noise Ratio (PSNR), Structural Similarity Index (SSIM), and Learned Perceptual Image Patch Similarity (LPIPS)~\cite{lpips}.
As in prior work \cite{barf}, because we perform camera optimization, to allow novel-view renders to be properly aligned, for a fair comparison we apply camera pose optimization to all methods before calculating the metrics.

\paragraph{Implementation details}
We implement our method with Nerfstudio~\cite{nerfstudio}.
We use the instant NGP~\cite{ingp} backbone with the default configurations and train all methods for 200k iterations to ensure convergence.
We empirically set $\lambda_{evs}{=}1$ for the event loss.
After training, 
to find the global embeddings, %
we perform 3k iterations of optimization.
When computing the metrics, we optimize the camera pose for 6k iterations. 
We choose $n=4$ in \cref{eq:blur_model} to make our RGB and event loss have a similar batch size of 597 and 588, respectively, which is the largest batch size that fits on our NVIDIA 3090 GPU with 24GB  VRAM.
Following Nerfstudio~\cite{nerfstudio}, we choose $D{=}32$ as our per-time embedding dimension; we ablate our choice in 
the supplement. %

\subsection{Results on our dataset}
\label{sec:ourdatares}

We show example qualitative results in \cref{fig:qualitative} and a quantitative summary in \cref{tab:quantitative}.
As shown in \cref{tab:quantitative}, our method with the embeddings and the learned mapper performs best.
It is interesting to note that our embedding strategy, combined with BADNeRF, already outperforms the state of the art, up to a degree where it outperforms methods that use events, although it uses only RGB data.
This highlights the importance of considering camera sensor modeling errors.
Still, our method of using events  
together with the per-time embeddings and the learned gamma mapping performs best. 

Besides the quantitative results, the differences between the reconstructions are more pronounced in \cref{fig:qualitative}.
With our method, sharper reconstructions are obtained.

\subsection{Results on the \evimo dataset}
\label{sec:evimodatares}

\begin{table*}
\centering
    \setlength{\tabcolsep}{4pt}
    \resizebox{\linewidth}{!}{%
\begin{tabular}{@{}l ccc ccc ccc ccc@{}}
\toprule
          & \multicolumn{3}{c}{\evimosceneA}  
          & \multicolumn{3}{c}{\evimosceneB} 
          & \multicolumn{3}{c}{\evimosceneC}  
          & \multicolumn{3}{c}{Average}   \\
\cmidrule(r){2-4} \cmidrule(r){5-7} \cmidrule(r){8-10} \cmidrule{11-13}
          & 
          \multicolumn{1}{c}{SSIM$\uparrow$} & \multicolumn{1}{c}{PSNR$\uparrow$} & \multicolumn{1}{c}{LPIPS$\downarrow$} &
          \multicolumn{1}{c}{SSIM$\uparrow$} & \multicolumn{1}{c}{PSNR$\uparrow$} & \multicolumn{1}{c}{LPIPS$\downarrow$} &
          \multicolumn{1}{c}{SSIM$\uparrow$} & \multicolumn{1}{c}{PSNR$\uparrow$} & \multicolumn{1}{c}{LPIPS$\downarrow$} &
          \multicolumn{1}{c}{SSIM$\uparrow$} & \multicolumn{1}{c}{PSNR$\uparrow$} & \multicolumn{1}{c}{LPIPS$\downarrow$} \\
\midrule
BADNeRF~\cite{BADNeRF}             
& 0.819 & 23.358 & 0.252 
& 0.786 & 20.726 & 0.396 
& 0.754 & 20.144 & 0.402 
& 0.786 & 21.410 & 0.350 \\
BADNeRF~\cite{BADNeRF} + Emb. 
& 0.852 & \second{24.912} & 0.217 
& \second{0.825} & 22.126 & \second{0.339 }
& 0.804 & 21.368 & 0.322 
& \second{0.827} & 22.802 & 0.292 \\

\midrule

Our method w/o Emb.
& \second{0.856} & 24.486 & \second{0.212 }
& 0.819 & \second{22.439} & 0.341 
& \second{0.805} & \second{21.619} & \second{0.321 }
& \second{0.827} & \second{22.848} & \second{0.291} \\

Our method      
& \best{0.880} & \best{26.152} & \best{0.177 }
& \best{0.855} & \best{24.247} & \best{0.288 }
& \best{0.839} & \best{22.927} & \best{0.257 }
& \best{0.858} & \best{24.442} & \best{0.241} \\
\bottomrule
\end{tabular}
}
\caption{\label{tab:evimo} 
{\bf Quantitative results for the \evimo dataset -- } 
We report the quantitative metrics for our method and the baselines on the test views.
Our method performs best.
}
\end{table*}

We also evaluate our method on \evimo. 
We report our results in \cref{tab:evimo}\footnote{E2NeRF~\cite{e2nerf} is excluded as the official implementation failed to reconstruct a clear scene despite our best efforts. 
This could be because the dataset is not front-facing, which requires careful mapping between the NeRF coordinate system and the world.
We tried various variants, including the exact same coordinate mapping as ours.
}. 
It is interesting to note that while our method performs best, once our per-time embedding strategy is used, RGB-only reconstruction with BADNeRF~\cite{BADNeRF} works almost as well as using events without per-time embeddings.
This further demonstrates the importance of incorporating these sensor modeling imperfections.

\subsection{Ablation study} %
\label{sec:ablation}

We now ablate our design choices.
For all our ablations, we use one indoor scene (`Dragon Max') and one outdoor scene (`Courtyard') from our dataset.

\begin{table}
  \centering
  \tabcolsep=8pt
  \resizebox{\linewidth}{!}{
  \begin{tabular}{@{}l ccc@{}}
    \toprule
    Method &  SSIM$\uparrow$ & PSNR$\uparrow$ & LPIPS$\downarrow$ \\
    \midrule
    BADNeRF~\cite{BADNeRF} & 0.769 & 23.474 & 0.424 \\
    BADNeRF~\cite{BADNeRF} + Zero Emb. & 0.776 & 23.440 & 0.403 \\
    BADNeRF~\cite{BADNeRF} + Our Emb. & \second{0.783} & 24.277 & 0.389 \\
    \midrule
    Ours $\mathcal{L}_{rgb}+\mathcal{L}_{evs}$ & \second{0.783} & \second{24.390} & \second{0.387} \\
    Ours $\mathcal{L}_{rgb}$ & \best{0.790} & \best{24.504} & \best{0.374} \\
    \bottomrule
  \end{tabular}
  }
  \caption{ 
  {\bf Ablation: per-time embedding -- }
  We report the quantitative metrics with various strategies of using per-time embeddings.
  Using our strategy to obtain the \emph{global} embedding significantly improves performance while simply training with embeddings and then using a zero embedding, which is the default strategy for NeRFStudio~\cite{nerfstudio} only shows a minor improvement.
  Furthermore, training only with the RGB loss, \cref{eq:rgb_loss_blurry}, to find the embeddings provides better results than using also the event loss.
  }
  \label{tab:ablation_embedding_loss}
\end{table}

\paragraph{Obtaining a good \emph{global} embedding is critical}
We first examine our learned per-time embeddings, which we already demonstrated its effectiveness in \cref{sec:ourdatares} and \cref{sec:evimodatares}, but this time, focusing on how one obtains the embeddings.
In \cref{tab:ablation_embedding_loss} we report the performance of BADNeRF~\cite{BADNeRF} with and without learned per-time embeddings, including our strategy of finding a global embedding that minimizes the RGB loss, and the typical default strategy of training with embeddings, which is to use a zero embedding.
As shown, our strategy significantly improves reconstruction quality, while the zero embedding offers only a minor improvement in SSIM and LPIPS, with negligible impact to PSNR.
Combining the event loss with the RGB loss is detrimental, likely because event data is sparse, whereas finding an embedding that accurately represents the scene requires considering all parts of the scene.
We further remind the reader that in Tables \ref{tab:quantitative} and
\ref{tab:evimo} 
our \emph{global} embeddings provide a significant boost in performance.

\begin{table}
  \centering
  \small
  \tabcolsep=12pt
  \resizebox{\linewidth}{!}{
  \begin{tabular}{@{}l ccc@{}}
    \toprule
    Method &  SSIM$\uparrow$ & PSNR$\uparrow$ & LPIPS$\downarrow$ \\
    \midrule
    Ours w/o events & 0.783 & 24.277 & 0.389 \\
    \midrule
    Ours + Linear mapping & 0.743 & 22.995 & 0.503 \\
    Ours + Normalized linear mapping & \second{0.783} & 24.302 & 0.400 \\
    Ours + MLP mapping~\cite{evDeblurNeRF} & \second{0.783} & \second{24.350} &\second{0.386} \\
    Ours + MLP mapping (both)~\cite{evDeblurNeRF} & 0.777 & 24.257 & 0.411 \\
    Ours + Gamma mapping & \best{0.790} & \best{24.504} & \best{0.374} \\
    \bottomrule
  \end{tabular}
  }
  \caption{
  {\bf Ablation: mapper -- }
  We report quantitative metrics for our gamma mapping (\cref{eq:gamma_model}), MLP mapping~\cite{evDeblurNeRF}, and simply no mapping (linear) with and without event normalization~\cite{enerf}.
  Using an MLP mapper on either events or both events and RGB helps, but it does not perform as well as our simple gamma mapper.
  }
  \label{tab:ablation_mapper}
\end{table}

\paragraph{Gamma mapping outperforms MLP mapping}
We report various mapping options for linking between RGB and event data in \cref{tab:ablation_mapper}.
Our learned gamma mapping provides a simple yet effective mapping strategy.
Using a simple linear mapping, that is, setting a constant default threshold for events as $w=0.2$ performs significantly worse.
Other mapping strategies, such as using an MLP %
\cite{evDeblurNeRF} or normalizing the events \cite{enerf} do not perform as well as ours.

\begin{table}
  \centering
  \tabcolsep=8pt
    \resizebox{\linewidth}{!}{
  \begin{tabular}{@{}lccc@{}}
    \toprule
    Method &  SSIM$\uparrow$ & PSNR$\uparrow$ & LPIPS$\downarrow$ \\
    \midrule
    BADNeRF~\cite{BADNeRF} w/o cam. opt. & 0.702 & 20.781 & 0.527 \\
    BADNeRF~\cite{BADNeRF} & \second{0.769} & \second{23.474} & \second{0.424} \\
    \midrule
    Our method w/o cam. opt. & 0.755 & 23.227 & 0.452 \\
    Our method &  \best{0.790} & \best{24.504} & \best{0.374} \\
    \bottomrule
  \end{tabular}
  }
  \caption{ 
  {\bf Ablation: camera optimization -- }
  We report quantitative metrics with and without camera optimization.
  Camera optimization is critical to achieving the best performance.
  }
  \label{tab:ablation_camera}
\end{table}

\paragraph{Camera optimization is important for blurry scenes}
We further look into the importance of camera optimization in \cref{tab:ablation_camera}.
As pointed out in previous work \cite{BADNeRF}, camera optimization is especially important when it comes to deblur settings.
For our data, this also holds.
Camera optimization during training significantly improves rendering quality.

\section{Conclusion and Future Work}

We introduced a NeRF framework that utilizes both RGB and event data to reconstruct a deblurred scene from data captured with fast camera motion.
Distinct from prior work, our cameras are separate, decoupled sensors
rather than aligned.
Central to our approach is learning
sensor imperfections through a data-driven manner using per-time embeddings, combined with a gamma mapping between RGB and event data.
To evaluate our method in this unique stereo capture setting, we introduced 
a new dataset, consisting of five indoor and outdoor captures each.
Empirically, we showed that our method outperforms the state of the art.

\paragraph{Limitations}
Our data collection process currently requires manual human annotation.
While this involves selecting three or more points to properly scale the scene to the calibrated extrinsics,
it remains a manual effort.  Additionally, our work relies on NeRF, while much of the field is transitioning 
to the 3D Gaussian splatting~\cite{gaussian-splat} framework.  Nonetheless, we believe our findings are \emph{backbone agnostic} and will also benefit 3D Gaussian splatting-based methods.

\clearpage

{
    \small
    \bibliographystyle{ieeenat_fullname}
    \bibliography{main}
}
\clearpage
\setcounter{page}{1}
\maketitlesupplementary

\section{Sensitivity analysis - embedding dimension}

We study the impact of the embedding dimension size in \cref{tab:ablation_emb_dim}.
As shown, the choice of the embedding dimension shows minor differences.
The differences are minor but our choice of $D{=}32$ provides the best overall results.
\begin{table}[h!]
  \centering
  \tabcolsep=12pt
   \resizebox{\linewidth}{!}{
  \begin{tabular}{@{}lccc@{}}
    \toprule
    Embedding dimension size &  SSIM$\uparrow$ & PSNR$\uparrow$ & LPIPS$\downarrow$ \\
    \midrule
    $D = 8$  & 0.785 & 24.216 & 0.382 \\
    $D = 16$ & \second{0.788} & \second{24.504} & 0.383 \\
    $D = 32$ & \best{0.790} & \second{24.504} & \best{0.374} \\
    $D = 64$ & 0.787 & \best{24.548} & \second{0.376} \\
    $D = 128$ & 0.785 & 24.500 & 0.386 \\
    \bottomrule
  \end{tabular}
 }
  \caption{ 
  {\bf Sensitivity analysis - embedding dimension -- }
  We report quantitative metrics for embedding dimensions.
  The differences are minor but our choice of $D{=}32$ provides the best overall results.
  }
  \label{tab:ablation_emb_dim}
\end{table}

\section{Detailed network architecture}

We use the Instant Neural Graphics Primitives (InstantNGP)~\cite{ingp} backbone with its default configuration.
We use 16 hashgrid levels, starting with a minimum resolution of $16\times16$, and with a maximum resolution of $2046\times2046$.
We use a hashmap size of 19, with each level having two feature dimensions.
We then use a Multi-Layer Perceptron (MLP) with two layers each with 64 neurons to convert hash encodings to a 16-dimensional feature, where 1 dimension represents density, and the rest is used as input to the color MLP.
For the color MLP head, we use three layers, again each with 64 neurons.

\section{Additional detail on data collection}

When collecting our data, we record our RGB data in a 12-bit High Dynamic Range (HDR) raw image to make the best use of our RGB sensor.
However, for ease of utilization and compatibility with existing non-HDR pipelines, we convert them to conventional RGB images.
Specifically, to convert an HDR image in $[0, 65535]$ to a conventional RGB image in $[0,255]$, we apply again gamma mapping, after clipping the dynamic range of the HDR sensor to its 95-th percentile to avoid focusing too much on saturated pixels, except for the `Engineer Building' and `House' sequences where we set it to 50-th and 70-th percentile, respectively---we use a different percentile because of the wide dynamic range of these two scenes due to shadows and the sun.
For the gamma mapping, we start from the standard value of 2.4, and lower or enhance its value until the scene looks natural.
We thus write:
\begin{equation}
    I_{RGB} = 
    255\times
    \min\left(
        1,
        \max\left(
        0,
        \left(\frac{I_{HDR}}{b}\right)^{\frac{1}{k}}
        \right)
    \right)
    .
\end{equation}
We provide the $b$ and $k$ values used for each scene in \cref{tab:scene_info}.
We further list the number of RGB images paired with event streams.
\begin{table}[h!]
  \centering
  \tabcolsep=12pt
   \resizebox{\linewidth}{!}{
  \begin{tabular}{@{}lccc@{}}
    \toprule
    Scenes &  $k$ & $b$ & Num Images \\
    \midrule
    Bag & 2.2 & 37937 & 378 \\
    Bicycle & 2.8 & 65487 & 372 \\
    Courtyard & 2.2 & 65458 & 442 \\
    Dragon Max & 1.0 & 48809 & 569 \\
    Engineer Building & 1.8 & 39550 & 547  \\
    House & 2.4 & 65523 & 408 \\
    Lab & 1.0 & 65529 & 569 \\
    Grad Lounge & 1.8 & 21169 & 369 \\
    Presentation Room & 1.7 & 12512 & 468\\
    Teddy Grass & 1.0 & 38158 & 569 \\
    \bottomrule
  \end{tabular}
 }
  \caption{ 
  {\bf Dataset statistics -- }
  We report the $k$ and $b$ values used to convert HDR images into RGB.
  We also report the number of frames associated with event streams.
  We set the gamma mapping value, $k$, by either enhancing or reducing the value manually to look natural, starting from 2.4.
  For the clipping value for HDR images, $b$, we set it to the 95-th percentile for each scene, except for `Engineer Building' and `House', which we set to the 50-th and 70-th percentile because of the wide dynamic range of these scenes.
  }
  \label{tab:scene-params}
  \label{tab:scene_info}
\end{table}

\section{Qualitative examples for \evimo}

In addition to the qualitative examples that we provide in the main paper, we show qualitative examples in \cref{fig:qualitative_evimo}.
As shown, our results provide the sharpest reconstructions.
It is interesting to note that, while in \cref{tab:evimo} both BADNeRF~\cite{BADNeRF} with our embeddings and Our method without embeddings provide similar results according to PSNR, using events provide qualitatively sharper reconstructions.
As most of the images are without detailed textures, this difference is not as pronounced in terms of quantitative metrics.

\def \evimowidth {0.19}

\begin{figure*}
    \centering
    \setlength{\tabcolsep}{2pt}
    \footnotesize
    \begin{tabular}{c c c c c c}
        & Ground Truth & \makecell{BADNeRF~\cite{BADNeRF}\\(RGB Only)} & \makecell{BADNeRF~\cite{BADNeRF} + Our Emb.\\(RGB Only)} & \makecell{Our Method w/o Emb\\(RGB + Events)} & \makecell{{\bf Our Method}\\(RGB + Events)} \\

        \rotatebox{90}{\hspace{0em} \evimosceneA } &
        {\includegraphics[width=\evimowidth\linewidth]{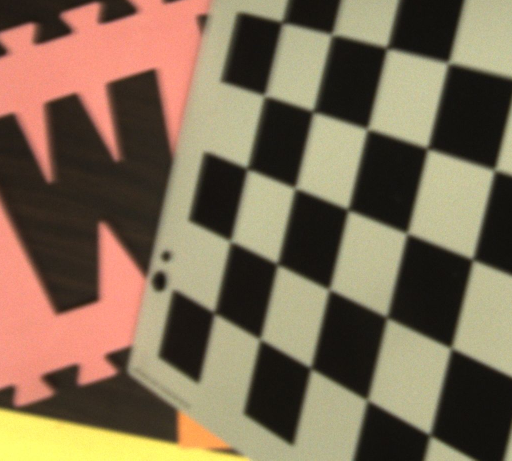}} &
        {\includegraphics[width=\evimowidth\linewidth]{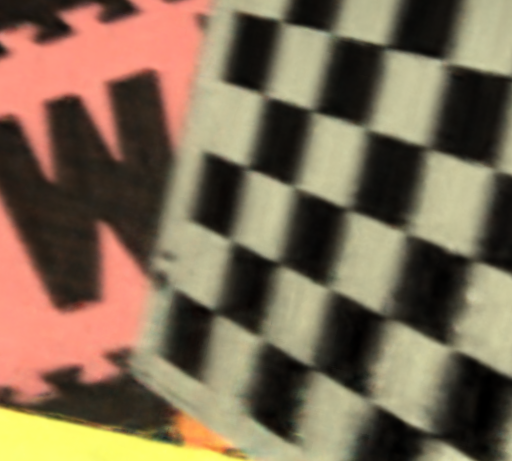}} &
        {\includegraphics[width=\evimowidth\linewidth]{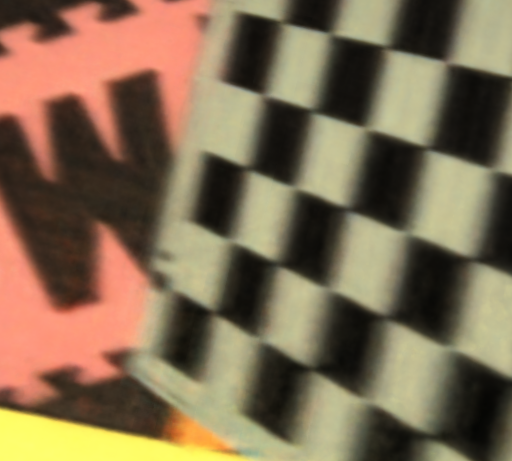}} & 
        {\includegraphics[width=\evimowidth\linewidth]{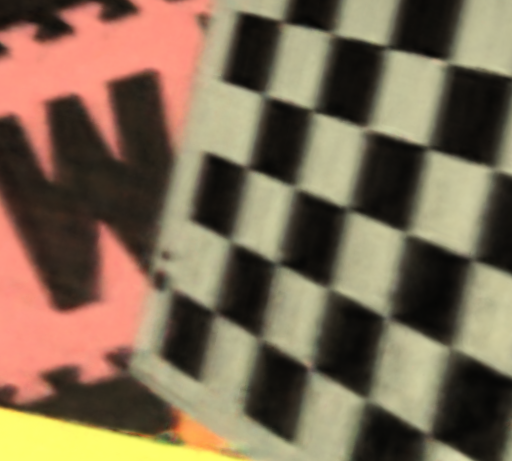}} &
        {\includegraphics[width=\evimowidth\linewidth]{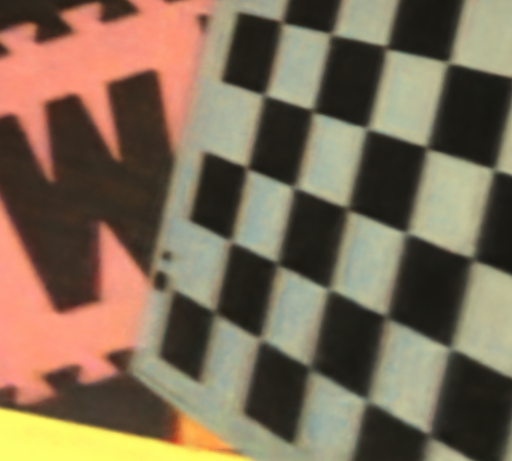}} \\

        \rotatebox{90}{\hspace{0em} \evimosceneB } &
        {\includegraphics[width=\evimowidth\linewidth]{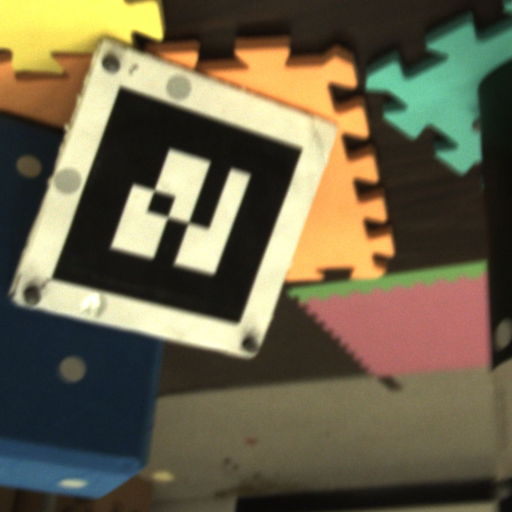}} &
        {\includegraphics[width=\evimowidth\linewidth]{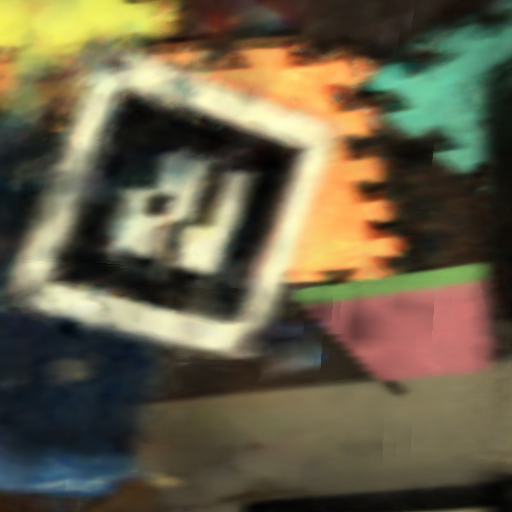}} &
        {\includegraphics[width=\evimowidth\linewidth]{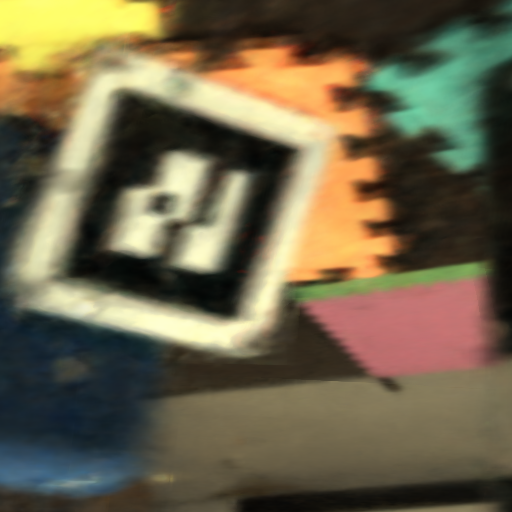}} & 
        {\includegraphics[width=\evimowidth\linewidth]{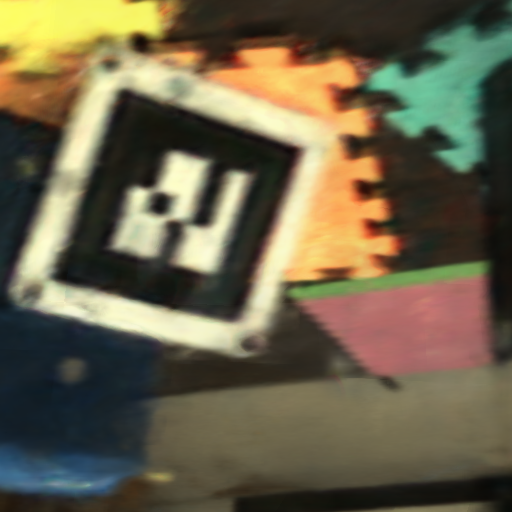}} &
        {\includegraphics[width=\evimowidth\linewidth]{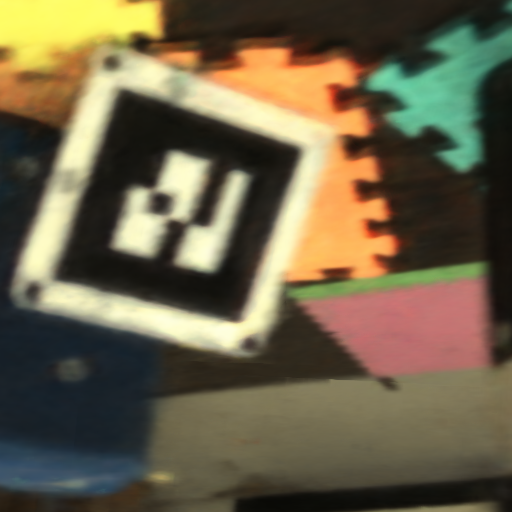}} \\

        \rotatebox{90}{\hspace{0em} \evimosceneB } &
        {\includegraphics[width=\evimowidth\linewidth]{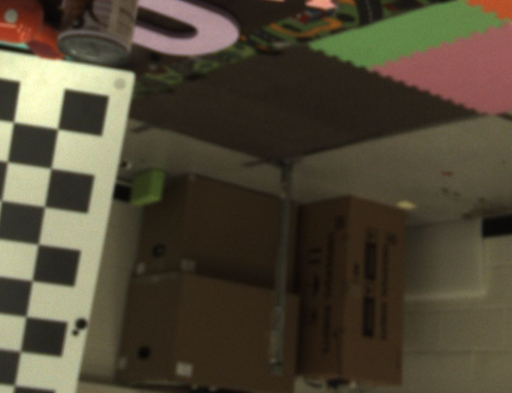}} &
        {\includegraphics[width=\evimowidth\linewidth]{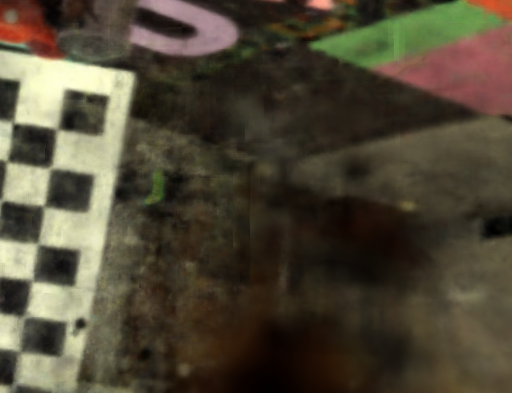}} &
        {\includegraphics[width=\evimowidth\linewidth]{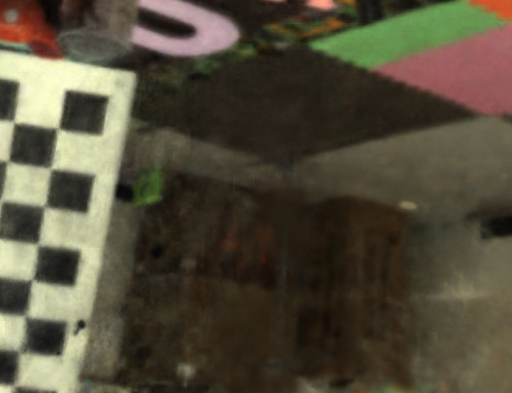}} & 
        {\includegraphics[width=\evimowidth\linewidth]{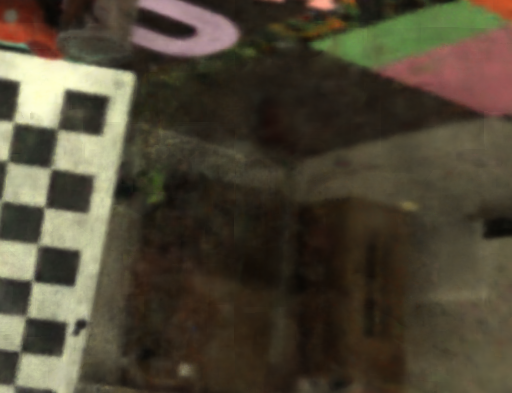}} &
        {\includegraphics[width=\evimowidth\linewidth]{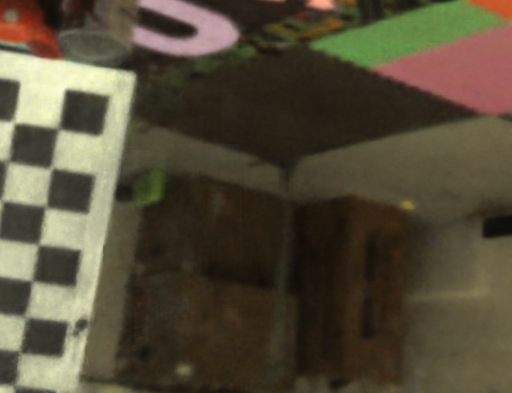}} \\

    \end{tabular}
    \caption{\textbf{\evimo qualitative examples -- } 
    We show qualitative examples of zoomed-in reconstruction cutouts from \evimo.
    As shown, our results provide the sharpest reconstructions. 
    }
    \label{fig:qualitative_evimo}
\end{figure*}

\end{document}